\documentclass{article} 
\usepackage{iclr2027_conference,times}
\usepackage{graphicx}
\usepackage{amsmath,amssymb}
\usepackage[section]{placeins}
\usepackage{comment}

\usepackage{hyperref}
\usepackage{pdflscape} 
\usepackage{url}
\hypersetup{hidelinks}

\title{When Does Synergy Help Active Feature \\ Acquisition? A PID-Based Study}

\author{Jie Li \\
Acutelines Data Science Group \\
University Medical Center Groningen \\
University of Groningen \\
Groningen, The Netherlands \\
\texttt{j.li04@umcg.nl} 
\And
Maruf A. Dhali \\
Department of Artificial Intelligence \\
University of Groningen \\
Groningen, The Netherlands \\
\texttt{m.a.dhali@rug.nl}
\And
Hjalmar R. Bouma \\
Acutelines Data Science Group\\
University Medical Center Groningen\\
University of Groningen\\
Groningen, The Netherlands \\
\texttt{h.r.bouma@umcg.nl}
}

\iclrfinalcopy 
\begin{document}

\maketitle

\begin{abstract}

Active feature acquisition (AFA) sequentially selects informative features under budget constraints. However, existing policies rarely distinguish whether information contributed by interacting features is redundant, unique, or synergistic. We introduce SynAFA, a state-dependent AFA policy that combines pairwise joint information and conditional information, with Partial Information Decomposition (PID) characterizing its information structure. Across five tabular datasets and MNIST-loop, performance is heterogeneous. SynAFA shows its strongest gains at low budgets on PhysioNet, where synergistic and redundant feature pairs are supported by permutation tests, but its advantage diminishes as budgets increase and does not depend on pair proposals. SynAFA performs significantly worse than nearly all baselines on MiniBooNE, and than CAE on MNIST-loop. Controlled synthetic experiments further show that, within budget, SynAFA's advantage rises as joint information becomes more synergy-dominated, including when total joint information is held approximately constant. Further analyses show that the budget-dependent erosion is not resolved by non-greedy local search, which improves a diagnostic set-level objective but leaves predictive performance no better, often significantly worse, and that improvements in this objective are weakly aligned with the fixed classifier's predictive utility. These findings characterize when pairwise synergy can benefit AFA while exposing a persistent challenge in translating local information into effective acquisition objectives.

\end{abstract}

\section{Introduction}
\label{sec:intro}
Active feature acquisition (AFA) selects measurements sequentially for each test instance, balancing predictive performance against acquisition cost. Laboratory tests, sensor queries, and survey items are costly, and a measurement's value depends on what is already known and what remains affordable. Existing approaches estimate conditional information \citep{covert2023learning,gadgil2024estimating}, learn generative surrogates \citep{ma2018eddi,zannone2019odin}, optimize acquisition policies \citep{shim2017pay}, or use acquisition-conditioned oracles \citep{valancius2023acquisition}.

A central difficulty is that two unobserved features may be informative together even when neither is useful alone. Meanwhile, individually informative features may carry redundant information. Partial Information Decomposition (PID) \citep{williams2010nonnegative} separates joint information into redundant, unique, and synergistic components. This distinction characterizes information structure, but does not determine whether an acquisition policy can exploit that structure under a budget.

We introduce SynAFA, which combines pairwise joint information and conditional information evaluated at the realized values of observed features, as a probe of when synergy helps rather than a claim of state-of-the-art performance. PID characterizes the information structure underlying its pair scores. The deployed policy uses total joint information, not separately weighted PID atoms. This study mainly addresses two questions: \emph{when does pairwise information with rich synergy help acquisition, and what limits its benefit as budgets grow?}

Our contributions are:
\begin{enumerate}
\item \textbf{Method:} a state-dependent policy that, at each step, compares single-feature proposals (scored by marginal information or by information conditional on already-observed values) with proposals for a not-yet-observed feature pair (scored by its joint information; the \emph{pair branch}) and acquires one feature from the best cost-adjusted proposal, with PID characterizing the information structure (\S\ref{sec:method}).
\item \textbf{When synergy helps:} controlled synergy--budget sweeps (varying synergy strength and budget), including fixed-information controls and a pair-branch ablation, identify benefits when the corresponding joint acquisition is feasible. On most tested real datasets, the pair branch contributes little (\S\ref{sec:when-synergy}).
\item \textbf{What limits the benefit as budgets grow:} non-greedy revision does not recover accuracy, and swap-level utility diagnostics and a duplicate-copy construction expose a limitation of aggregating local pairwise scores into a set objective (\S\ref{sec:analysis}).
\end{enumerate}

\section{Related Work and Problem Formulation}
\label{sec:related}
\paragraph{Active feature acquisition.}
Greedy methods estimate the conditional value of the next feature \citep{covert2023learning,gadgil2024estimating}. Generative methods infer unobserved measurements to evaluate candidate acquisitions \citep{ma2018eddi,zannone2019odin}. Reinforcement learning (RL) methods optimize sequential acquisition and stopping \citep{shim2017pay,mnih2015human}. Acquisition-conditioned oracles \citep{valancius2023acquisition} and information templates \citep{huang2025information} explicitly address jointly informative acquisitions. Static methods such as concrete autoencoders (CAE) instead learn a population-wide subset against a downstream objective \citep{balin2019concrete}. SynAFA complements these approaches by relating acquisition outcomes to the redundant, unique, and synergistic composition of pairwise information. We compare policies within AFABench \citep{schutz2026afabench}, and see also \citet{aronsson2025survey}.

\paragraph{Problem and notation.}
For features $X_1,\ldots,X_D$, binary target $Y$, and costs $c_j$, the observed state is $(S,x_S)$, where $S$ is the acquired set. A policy chooses an unobserved feature or stops to predict, subject to a hard budget $B$, so $\sum_{j\in S}c_j\le B$. Information tables are estimated from binarized training features. The test-time policy conditions on observed feature values. Under evaluation, a shared external classifier predicts using the acquired subset. Detailed preprocessing and the hard-budget protocol are in Appendices~\ref{app:preprocessing-protocol} and \ref{app:external-classifier}.

\paragraph{Pairwise information and PID.}
For two sources, PID decomposes their joint mutual information as
\begin{equation}
I(X_j,X_k;Y)=R+U_j+U_k+\mathrm{Syn},
\label{eq:pid-main}
\end{equation}
where $R$ is shared redundant information, $U_j,U_k$ are source-unique information, and $\mathrm{Syn}$ is synergistic information available only jointly. We use the Williams--Beer $I_{\min}$ definition \citep{williams2010nonnegative} (see Appendix~\ref{app:pid-definitions}). PID has also been used to characterize static feature selection \citep{wollstadt2023rigorous}. Our focus is sequential acquisition with realized-value conditioning: $I(Y;X_j\mid X_k=x_k)$ can vary with the observed value $x_k$, unlike a population-level pair score.

\section{SynAFA}
\label{sec:method}
\subsection{Offline information quantities}
\label{sec:method-v1}
Using the training data, we precompute
\begin{equation}
M_j=I(X_j;Y),\qquad V_{jk}=I(X_j,X_k;Y),\qquad
C_{j\mid k,v}=I(Y;X_j\mid X_k=v).
\label{eq:offline-tables}
\end{equation}
These correspond to \texttt{single\_value}, \texttt{pair\_value}, and \texttt{cond\_single\_value}. PID decomposes $V_{jk}$ for characterization and validation. Individual atoms are not directly used in the deployed acquisition score.

The conditional table stores $I(Y;X_j\mid X_k=v)$ separately for each observed value $v$ of $X_k$, rather than only its average over $v$:
\begin{equation}
I(Y;X_j\mid X_k)=\sum_v p(X_k=v)\,C_{j\mid k,v}.
\label{eq:conditional-average}
\end{equation}
Different observed values can therefore produce different candidate scores. Conditioning uses one observed feature at a time; the maximum below does not estimate full-state information $I(Y;X_j\mid X_S=x_S)$.

\subsection{State-dependent acquisition}
Given observed state $(S,x_S)$, each unobserved feature receives
\begin{equation}
s_j=\max\left\{M_j,\max_{k\in S}C_{j\mid k,x_k}\right\},
\label{eq:single-score}
\end{equation}
with $s_j=M_j$ when $S$ is empty. The policy compares single-feature proposals $s_j-\lambda c_j$ with pair proposals $V_{jk}-\lambda(c_j+c_k)$, where $j,k\notin S$. If a pair wins, only its member with larger marginal information is acquired at that step. Its partner can subsequently become valuable through realized-value conditioning, while completion of the pair is not guaranteed. Acquisition stops when no cost-adjusted proposal is positive, subject to the hard-budget evaluation procedure in Appendix~\ref{app:preprocessing-protocol}.

The deployed pair branch does \emph{not} additionally require the full pair cost to fit within the remaining budget. Although the acquired feature must consider the hard budget, this rule can start a pair that the budget cannot afford to complete. We retain this rule for the primary comparisons and test a feasibility mask as a post hoc mechanism diagnostic (\S\ref{sec:feasibility}). The policy is one-step-ahead and does not optimize a complete acquisition sequence.

\paragraph{Why compare single and pair proposals?}
For two independent fair bits with an XOR target, each bit alone carries no information about the target, but together they determine it (one bit). Single-feature scores would never select either bit, whereas a pair proposal can start the acquisition. Once one bit is observed, the conditional table makes the other informative. Costs and competing features can still prevent the pair from being completed.

\begin{figure}[t]
\centering
\small
\fbox{\begin{minipage}{0.94\linewidth}
\textbf{Information structure:} $V_{jk}=R+U_j+U_k+\mathrm{Syn}$ (PID characterization).\\[3pt]
\textbf{Offline:} training data $\longrightarrow (M_j,V_{jk},C_{j\mid k,v})$.\\[3pt]
\textbf{Online:} observed state $(S,x_S)$ $\longrightarrow$ compare single/pair proposals
$\longrightarrow$ acquire one feature $\longrightarrow$ update state.\\[3pt]
\textbf{Prediction:} acquired values and mask $\longrightarrow$ shared external classifier.
\end{minipage}}
\caption{SynAFA in one schematic. PID characterizes joint information; test-time decisions use the three precomputed information tables. Pair proposals initiate one acquisition rather than committing to both features.}
\label{fig:synafa-overview}
\end{figure}

\subsection{Scaling to high dimensions: an amortized estimator (SynAFA v2)}
\label{sec:method-v2}
Exact pairwise estimation needs $O(D^2)$ pair evaluations, each over $O(n)$ training samples, which is feasible for moderate $D$ but expensive at image scale ($D=784$, yielding $\binom{784}{2}=306{,}936$ pairs). To scale SynAFA to this setting, SynAFA v2 trains an amortized estimator that maps a pair's empirical frequency statistics to its information values, including $V_{jk}$ from one fixed sample of the whole training set, and $C_{j\mid k,v}$ from a fixed sample conditioned on $X_k=v$. This computes each pair's, or conditioned variable's, frequency counts at once, rather than estimating them exactly from the full training set. SynAFA v2 is therefore the variant of SynAFA designed for high-dimensional scaling, rather than a different policy. Estimator details and controls are in Appendix~\ref{app:amortized}.

\section{Experimental Setup and Broad Benchmark}
\label{sec:experiments}
\paragraph{Evaluation protocol.}
We evaluate six datasets: five tabular datasets (CKD, ACTG175, BankMarketing, PhysioNet \citep{silva2012predicting}, and MiniBooNE), and one imaging dataset, MNIST-loop ($D=784$), a binary relabeling of MNIST \citep{lecun1998gradient}. Dataset dimensions and test sizes are in Table~\ref{tab:dataset-overview}. Hard budgets for the five tabular datasets follow AFABench's default configuration \citep{schutz2026afabench}, and MNIST-loop's budgets (5, 10, 20) and the synthetic sweep's budgets (3, 5, 8) were fixed for this paper and not tuned on validation or test performance. Comparisons within each dataset use the same hard budget and the same externally trained masked MLP. This controls the prediction model across policies, but does not guarantee equal generalization across their mask distributions. Real-data costs are uniform, while the controlled sweep uses heterogeneous costs. SynAFA uses $\lambda=0.01$ in the tabular and synthetic hard-budget experiments, without dataset-specific tuning. ForMNIST-loop (SynAFA v2), $\lambda=0.001$ was fixed because pixel features are sparse, and a post hoc sensitivity analysis (Appendix~\ref{app:lambda-mnist}) finds it the best of three values tested.

For paired test instances, $\Delta\mathrm{Acc}=n^{-1}\sum_i[\mathbf{1}(\hat y_i^{\rm SynAFA}=y_i)-\mathbf{1}(\hat y_i^{\rm baseline}=y_i)]$. Accuracy lies below the majority-class rate on PhysioNet and BankMarketing (Table~\ref{tab:dataset-overview}), so only paired differences are interpreted. Significance uses a two-sided paired bootstrap with 10,000 resamples.  Tests are per comparison and unadjusted for multiplicity. All results use the corrected hard-budget harness (Appendix~\ref{app:full-tables}). With heterogeneous costs, a common budget is a cost ceiling, not a guarantee of equal feature counts.

The eight shared tabular baselines cover oracle-based acquisition (AACO), permutation importance, static selection (CAE), greedy methods (GDFS, DIME), and RL (JAFA and OL with/without an observation mask). EDDI and ODIN are additionally evaluated on the two smallest datasets.  Appendix~\ref{app:full-tables} reports numerical comparisons for PhysioNet and MiniBooNE against all eight baselines, together with the available comparison summaries for the remaining datasets.

\paragraph{Heterogeneous performance.}
CKD and ACTG175 show mostly non-significant differences across budgets. BankMarketing is mixed and non-monotonic. PhysioNet gives the strongest positive tabular result at the lowest budget, but the advantage diminishes as the budget increases. MiniBooNE is a strong negative case, with significant losses in 22 of 24 budget--baseline comparisons. Thus the broad benchmark motivates a conditional explanation of success, rather than a claim of uniform superiority (Tables~\ref{tab:v1-headline} and \ref{tab:physionet-clean}).

\begin{table}[htbp]
\centering
\caption{Results of the tabular benchmark: accuracy differences at the lowest and highest tested budgets, against the adaptive AACO and static CAE. Positive values favor SynAFA. $^*p<0.05$ without multiplicity correction. Endpoints summarize scope, not monotonicity, and all intermediate budgets are in Table~\ref{tab:v1-headline}. CKD ($n=80$) stars are bootstrap-based and are not significant under an exact sign test (Table~\ref{tab:ckd-actg}). Small CKD and ACTG175 test sets ($n=80$ and $429$) limit precision, non-significant differences thus do not establish equivalence.
}
\label{tab:broad-endpoints}
\small
\setlength{\tabcolsep}{4pt}
\begin{tabular}{lcrrrr}
\hline
 & Budgets & \multicolumn{2}{c}{vs. AACO} & \multicolumn{2}{c}{vs. CAE} \\
Dataset & low/high & Low & High & Low & High \\
\hline
CKD & 2/7 & $-0.050^*$ & $+0.000$ & $+0.038$ & $+0.025$ \\
ACTG175 & 3/10 & $-0.023$ & $-0.009$ & $+0.016$ & $+0.009$ \\
BankMarketing & 2/7 & $+0.034^*$ & $-0.068^*$ & $+0.000$ & $-0.045^*$ \\
PhysioNet & 5/15 & $+0.011$ & $-0.038^*$ & $+0.037^*$ & $-0.005$ \\
MiniBooNE & 5/15 & $-0.034^*$ & $-0.033^*$ & $-0.025^*$ & $-0.044^*$ \\
\hline
\end{tabular}
\end{table}

\begin{table}[htbp]
\centering
\caption{PhysioNet: mean accuracy difference against eight baselines and counts of significant wins/ties/losses at each budget ($n=2{,}400$). Ties denote non-significant differences, not equivalence. Tests are unadjusted across comparisons. Full effects are in Table~\ref{tab:physionet-clean}.}
\label{tab:physionet-summary}
\small
\begin{tabular}{crccc}
\hline
Budget & Mean $\Delta\mathrm{Acc}$ & Wins & Ties & Losses \\
\hline
5 & $+0.026$ & 6 & 2 & 0 \\
10 & $+0.005$ & 2 & 5 & 1 \\
15 & $-0.011$ & 1 & 5 & 2 \\
\hline
\end{tabular}
\end{table}

\section{When Does Synergy Help?}
\label{sec:when-synergy}
The benchmark raises a question about conditions for success. We first examine whether the strongest positive case contains synergistic pairwise structure supported by permutation tests. As real data doesn't allow us to change this structure, we use controlled sweeps to investigate whether the benefit comes from the information itself or from the acquisition feasibility.

\subsection{Real-data validation: PhysioNet}
\label{sec:physionet}
Stratified permutation tests support synergistic and redundant feature structure in PhysioNet. In a scan of the 351 testable pairs (of 378 usable) with Benjamini--Hochberg FDR correction, GCS/PaO$_2$ was the strongest synergy pair ($\mathrm{Syn}=0.0127$ bits, raw permutation $p=0.0005$, the smallest value obtained from 2{,}000 permutations). The pair survived a discovery/held-out-validation procedure. BUN/creatinine, tested separately as a clinically motivated candidate, supplies a confirmed redundant pair. The scan uses the full cohort, including the 2{,}400 evaluation patients, while SynAFA never uses its output. Procedures and a separate illustrative example (PaO$_2$/FiO$_2$) are in Appendix~\ref{app:physionet-validation}.

In the held-out check, the scan in the discovery half (6{,}000 patients per half) selected 51 of 351 pairs after FDR correction. Only these pairs were tested on the validation half, with correction across that confirmatory set. 36 of the 51 (70.6\%) replicated, including GCS/PaO$_2$. This separates discovery from confirmation within that check.

This structure coincides with SynAFA's advantage at low budgets(Table~\ref{tab:physionet-summary}). However, removing the pair branch does not change that advantage (Appendix~\ref{app:pair-off}), indicating that the advantage is not coming from pair proposals. The controlled experiments below vary the designated pair's information structure while holding the remaining construction fixed.

\subsection{Controlled synergy--budget sweep}
\label{sec:dose-response}
We construct a binary dataset with $D=24$, including a designated pair S1 (cost 1) and S2 (cost 5), and 22 independent background features of decreasing informativeness. The designated pair's contribution to the outcome logit is
\begin{equation}
(1-\alpha)[w_1(2S_1-1)+w_2(2S_2-1)]
+\alpha w_{\rm xor}[2(S_1\oplus S_2)-1],
\label{eq:synthetic-logit}
\end{equation}
where $w_1=w_2=0.7$, $w_{\rm xor}=1.4$, and $\alpha\in\{0,0.25,0.5,0.75,1\}$. The pair changes from additive to pure XOR structure. SynAFA and independently trained AACO, CAE, and permutation baselines use the same pipeline at budgets 3, 5, and 8, with 6,000 test instances per comparison and dataset instance (five independently generated dataset instances per $\alpha$; Appendix~\ref{app:synthetic-generative}).

\begin{figure}[t]
\centering
\includegraphics[width=\linewidth]{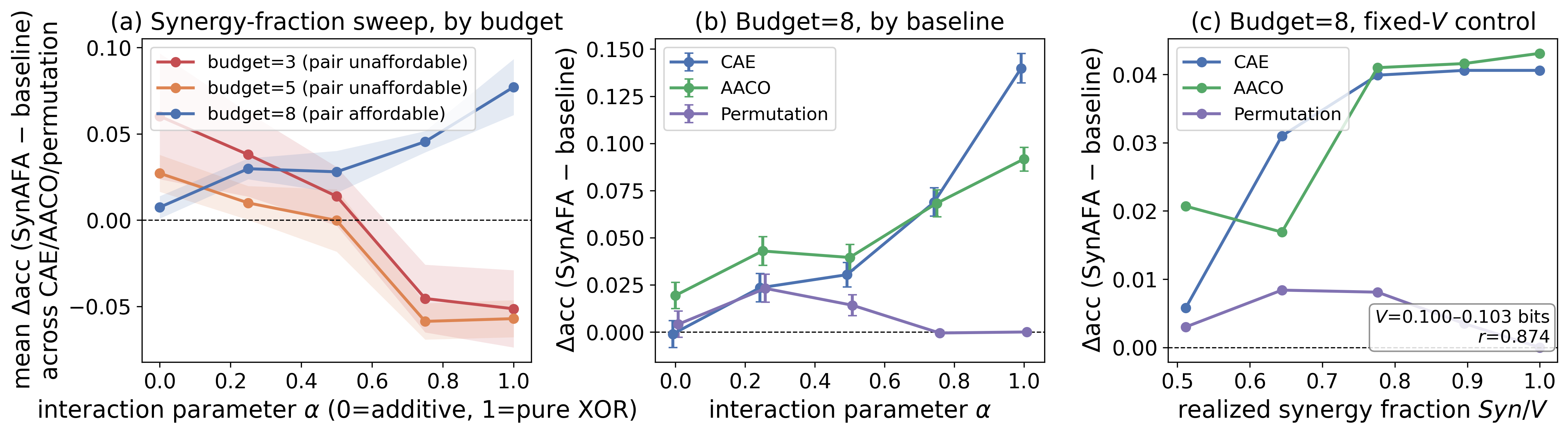}
\caption{Controlled synergy sweeps. (a) Mean SynAFA advantage by budget (five dataset instances; band: $\pm1$ SD across instances); the designated pair costs 6, exceeding budgets 3 and 5 but fitting budget 8. (b) Individual baseline comparisons at budget 8, pooled over instances (95\% paired-bootstrap intervals). (c) Fixed-$V$ control: advantage versus realized synergy fraction with joint information approximately constant. Full results and calibration are in Appendix~\ref{app:full-tables}.}
\label{fig:synergy-dose-response}
\end{figure}

Figure~\ref{fig:synergy-dose-response} (a) and (b) show the budget-dependent trend for the deployed policy. The pair costs 6, so it is unaffordable at budgets 3 and 5 and affordable at budget 8. At budget 8, SynAFA's mean advantage (five dataset instances) rises from $+0.007$ at $\alpha=0$ to $+0.077$ at $\alpha=1$ (instance-bootstrap 95\% interval $[0.063,0.089]$), reaching $+0.140$ against CAE and $+0.092$ against AACO, with permutation roughly tied. The descriptive correlation across the five prescribed levels is $r=+0.946$. At budgets 3 and 5, the correlations are $-0.974$ and $-0.945$. Each correlation has the same sign in all five dataset instances. Full matrices are in Tables~\ref{tab:synergy-dose-response} and \ref{tab:synergy-dose-response-full}. Removing the pair branch reverses the trend at budget 8 (advantage falls from $+0.006$ to $-0.043$; Table~\ref{tab:pair-off-alpha}), so the rise is due to pair proposals (Appendix~\ref{app:pair-off}).

The positive trend at budget 8 is not strictly monotonic and is strongest against CAE and AACO, while permutation returns to no advantage at high $\alpha$. Moreover, $\alpha$ changes information composition without holding total information fixed. Under this symmetric $I_{\min}$ construction, $U_a=U_b=0$, and $\mathrm{Syn}/V$ rises from 0.52 to 1.00 even though neither $\mathrm{Syn}$ nor $V$ is monotonic (Table~\ref{tab:synergy-pid-decomposition}). The experiment therefore concerns increasingly synergy-dominated information, not an isolated change to one PID atom. 

\subsection{Fixed-information control}
\label{sec:fixed-v}
There is still another question: does the benefit at budget 8 reflect information composition or simply more joint information? To address this, a second sweep recalibrates the additive and XOR weights for target fractions $\mathrm{Syn}/V\in\{0.52,0.65,0.78,0.90,1.00\}$ while holding $V$ approximately fixed. Realized $V$ is 0.1003--0.1032 bits and realized $\mathrm{Syn}/V$ is 0.512--1.000 (Table~\ref{tab:fixed-v-calibration}). This construction cannot be reliably calibrated for synergy fraction below about 0.5, so the control covers only the upper half of the range, roughly 0.5 to 1, not the full 0 to 1.

Figure~\ref{fig:synergy-dose-response} (c) shows that mean advantage rises from $+0.010$ to $+0.028$ with synergy fraction ($r=0.874$) at budget 8. The largest-fraction effects are $+0.043$ against AACO and $+0.041$ against CAE, while permutation remains nearly no advantage (Table~\ref{tab:fixed-v-full-significance}). This trend cannot be explained simply by an increase in total pair information. Together, the controls suggest that the benefit comes from synergy-dominated pairwise information within the tested constructions when the joint acquisition is feasible. Ablating the pair branch shows that this benefit is concentrated at high synergy fractions and comes from pair proposals (Appendix~\ref{app:pair-off}).

To check whether this trend replicates across datasets, we pooled five separately generated dataset instances, each with its own trained policies, into 30,000 paired test comparisons. Within each instance, the correlation between synergy fraction and mean advantage is positive, and each instance improves from the lowest fraction to the highest (Table~\ref{tab:fixed-v-full-significance}). The significance markers in the pooled tables only reflect randomness in which patients are included in the test sample. They do not test whether the trend remains across these dataset instances.

\subsection{Mechanism: can the policy complete the pair?}
\label{sec:feasibility}
Looking at what acquisition records explain the tight-budget reversal. At budget 3, increasing $\alpha$ causes SynAFA to abandon cheap S1 (in four of five dataset instances at $\alpha=1$), because S1 carries no information by itself under pure XOR. At budget 5, it acquires expensive S2 alone (four of five instances at $\alpha=1$). At budget 8, both features are acquired in 100\% of test instances in all dataset instances at $\alpha\ge0.75$, whereas completion at $\alpha=0$ is unreliable (0--100\%, mean 30\%; Table~\ref{tab:synergy-acquisition-rates}). Affordability alone therefore does not guarantee that SynAFA completes the pair.

We repeat the sweep with one change that is masking pair proposals whose combined cost exceeds the remaining budget. The mask does not change the results at budget 8. At budgets 3 and 5 it removes the negative mean relative effects at high synergy fractions. Table~\ref{tab:feasibility-main} gives the $\alpha=1$ case, and full sweeps are in Tables~\ref{tab:synergy-budget-mask-sensitivity} and \ref{tab:synergy-budget-mask-sensitivity-full}. This intervention supports a feasibility limitation of the deployed proposal rule as an explanation for the high-synergy losses, rather than saying that synergy is generally harmful under tight budgets. This mask is a post-hoc diagnostic used only here, and primary results retain the deployed rule.

\begin{table}[htbp]
\centering
\caption{Feasibility diagnostic at pure XOR ($\alpha=1$): mean accuracy advantage over AACO, CAE, and permutation, averaged over five dataset instances. Masking restores positive relative performance when the pair is unaffordable (budgets 3 and 5: the mean advantage is negative before and positive after masking in all five dataset instances). Budget 8 is unchanged.}
\label{tab:feasibility-main}
\small
\begin{tabular}{crr}
\hline
Budget & Deployed & Feasibility-masked \\
\hline
3 & $-0.051$ & $+0.064$ \\
5 & $-0.057$ & $+0.063$ \\
8 & $+0.077$ & $+0.077$ \\
\hline
\end{tabular}
\end{table}

\section{What Explains the Erosion of the Benefit?}
\label{sec:analysis}
On real data SynAFA's relative performance worsens as budgets grow (PhysioNet, MNIST-loop), and under uniform costs feasibility alone doesn't explain this. One natural explanation is irreversible greedy selection, in which a feature that scores well at the current step may make later features redundant, and a feature that scores poorly at the current step may only become valuable once another feature is acquired first. Greedy scoring cannot see either possibility in advance, and the choice cannot be reversed once made. We test this explanation by revising acquired sets, then examine whether the diagnostic objective is aligned with prediction.

\subsection{MNIST-loop stress test and budget erosion}
\label{sec:mnist_loop}
\label{sec:pattern}
MNIST-loop relabels digits $\{0,6,8,9\}$ against $\{1,2,3,4,5,7\}$, providing an imaging task in which spatially separated pixels can carry joint information. SynAFA v2 significantly outperforms AACO at all tested budgets but loses to CAE, with the gap widening from budget 5 to 20 (Table~\ref{tab:mnist-summary}).

\begin{table}[htbp]
\centering
\caption{MNIST-loop: accuracy(SynAFA v2)$-$accuracy(baseline), evaluated with the shared classifier ($n=12{,}000$). $^*$ denotes an unadjusted paired-bootstrap $p<0.05$.}
\label{tab:mnist-summary}
\small
\begin{tabular}{lccc}
\hline
Baseline & Budget 5 & Budget 10 & Budget 20 \\
\hline
AACO & $+0.127^*$ & $+0.149^*$ & $+0.070^*$ \\
CAE & $-0.023^*$ & $-0.034^*$ & $-0.070^*$ \\
Permutation & $+0.020^*$ & $+0.001$ & $-0.018^*$ \\
\hline
\end{tabular}
\end{table}

Separate controls show that the hard-budget fallback is never invoked (all 12,000 test instances reach each budget by the policy's own choice), and that the gain over the static ablations is due to the realized value adaptation. Estimator controls indicate a contribution from amortization error that increasing the precompute sample shrinks the accuracy gap from 0.0128 to 0.0005 on a reduced $D=90$ subset. Under exact scoring at $D=90$, local search improves accuracy at budgets 5 and 20 but not 10. This comparison changes dimensionality and estimator together and does not show whether the aggregation objective is aligned with predictive accuracy (Appendices~\ref{app:mnist-diagnostics} and \ref{app:amortized}). Therefore, the erosion on PhysioNet and MNIST-loop motivates testing whether revising irreversible greedy selections recovers predictive utility.

\subsection{Non-greedy revision of a diagnostic objective}

\label{sec:local-search-test}
For a fixed instance, define
\begin{equation}
T(S)=\sum_{j\in S}\max\left\{M_j,\max_{k\in S\setminus\{j\}}C_{j\mid k,x_k}\right\},
\label{eq:diagnostic-objective}
\end{equation}
where a singleton contributes $M_j$. $T(S)$ is a diagnostic aggregation of local scores, not true set-level mutual information. SynAFA itself does not define a global set objective, so this search is not an exact non-greedy optimization of the deployed policy. Starting from each instance's acquired set, the corrected search evaluates replacements by fully recomputing $T(S')$, accepting the best swap only if $T(S')>T(S)$. Every accepted move therefore improves the diagnostic objective; all instances converged within 200 rounds. This is a retrospective diagnostic of revisability, not a deployable acquisition rule that refunds previous measurements. Full methodology is in Appendix~\ref{app:local-search}.

Across nine settings, which are three budgets each for MNIST-loop ($D=784$, amortized), PhysioNet ($D=41$, exact), and MiniBooNE ($D=50$, exact), local search yields no significant accuracy improvement and seven significant decreases (Table~\ref{tab:failure-summary}; $p$-values in Table~\ref{tab:local-search-generalization}). Since increasing $T(S)$ does not recover accuracy, we next examine the predictive utility of accepted swaps.

\subsection{Do objective improvements predict utility improvements?}
\label{sec:objective-alignment}
For each accepted swap, we measure $\Delta T$ and $\Delta U$, where $U$ is the fixed external classifier's predicted probability of the true class. A random-swap control uses the same removal and a uniformly random replacement. Although $\Delta T>0$ by construction, mean $\Delta U$ is small in every setting (from $-0.018$ to $+0.0014$) and negative in six of nine. Pearson correlations range from $-0.068$ to $+0.107$; 0.419--0.533 of accepted swaps increase utility, with setting-dependent differences from random replacements (Table~\ref{tab:objective-alignment}; all results use the full test sets).

\begin{table}[htbp]
\centering
\caption{Non-greedy revision under the diagnostic objective, full test sets. Accuracy changes compare revised and original sets; mean utility changes and correlations use accepted swaps. These point estimates summarize Tables~\ref{tab:local-search-generalization} and \ref{tab:objective-alignment}; $^*$ denotes unadjusted $p<0.05$ for accuracy.}
\label{tab:failure-summary}
\small
\begin{tabular}{lrrrr}
\hline
Dataset & Budget & $\Delta\mathrm{Acc}$ & Mean $\Delta U$ & $\rho(\Delta T,\Delta U)$ \\
\hline
MNIST-loop & 5 & $-0.0025$ & $+0.0014$ & $+0.073$ \\
MNIST-loop & 10 & $-0.0213^{*}$ & $+0.0007$ & $+0.107$ \\
MNIST-loop & 20 & $-0.0322^{*}$ & $-0.0024$ & $+0.044$ \\
PhysioNet & 5 & $-0.0371^{*}$ & $-0.0116$ & $-0.010$ \\
PhysioNet & 10 & $-0.0150^{*}$ & $-0.0022$ & $-0.068$ \\
PhysioNet & 15 & $-0.0042$ & $+0.0004$ & $-0.048$ \\
MiniBooNE & 5 & $-0.0404^{*}$ & $-0.0180$ & $+0.095$ \\
MiniBooNE & 10 & $-0.0148^{*}$ & $-0.0038$ & $+0.060$ \\
MiniBooNE & 15 & $-0.0065^{*}$ & $-0.0039$ & $+0.006$ \\
\hline
\end{tabular}
\end{table}

\paragraph{What the alignment diagnostic measures.}
These correlations are conditional on a swap being selected and accepted, so $\Delta T>0$ throughout. They describe the relationship between improvement magnitudes along this search, not ranking accuracy over all candidate swaps. Moreover, swaps from the same instance are dependent, so the reported confidence intervals use an instance-level cluster bootstrap and the correlations are treated as descriptive.

On MNIST-loop, selected swaps improve utility more often than random replacements (e.g., 0.506 versus 0.400 at budget 5, a 10.6-percentage-point difference), yet their mean $\Delta U$ is only $+0.0014$ there and $-0.0024$ at budget 20 (0.485 versus 0.418). Improvement frequency therefore does not imply a positive average gain, nor do these results establish universal equivalence to random selection. In MiniBooNE at budgets 10 and 15, selected swaps improve utility slightly less often than random ones.

The pattern occurs with both exact and amortized scores. Both $\Delta U$ and accuracy depend on the external classifier, and these diagnostics do not establish misalignment independent of the classifier. Missing higher-order interactions remain possible, but the observations directly implicate the alignment of the tested objective rather than higher-order information as the primary cause.

The following construction shows that overcounting can occur without estimation error or higher-order synergy.
Let $M\ge2$ and $X_1=\cdots=X_M=Z$ be identical copies with $I(Z;Y)=I_0>0$, and $S=\{1,\ldots,M\}$. Then
\begin{equation}
T(S)=MI_0,\qquad I(X_S;Y)=I_0,\qquad T(S)/I(X_S;Y)=M.
\label{eq:redundancy-overcounting}
\end{equation}
Given any copy $X_k=x_k$ with positive probability, every other copy $X_j$ is forced to equal $x_k$. $X_j$ therefore becomes deterministic and carries no remaining information about $Y$, yielding $C_{j\mid k,x_k}=0$. Each term in Eq.~\ref{eq:diagnostic-objective} therefore reduces to $\max(I_0,0)=I_0$, whereas observing all copies together reveals only the information in $Z$, namely $I_0$. 

The marginal information floor fails to discount acquired redundancy. By the chain rule, true joint information has zero increments after the first identical copy, while $T(S)$ counts $M_j$ for every copy, which causes a structural failure. It doesn't mean that redundancy explains every bad swap, or that all sets are overestimated by size.

\section{Discussion and Limitations}
\label{sec:discussion}
Across the six real datasets and the controlled synthetic family, this paper studies one question through three stages: \emph{when does synergy-rich pairwise information help acquisition, and what limits that benefit as budgets grow?} The broad benchmark in \S\ref{sec:experiments} shows that SynAFA's advantage is heterogeneous, motivating a search for the conditions under which it succeeds. \S\ref{sec:when-synergy} located one such condition on PhysioNet, where permutation tests confirm synergistic and redundant pairwise structure. Controlled sweeps show that the advantage rises with increasing synergy fraction when the pair can be acquired within budget, though PhysioNet's own advantage does not depend on the pair branch. \S\ref{sec:analysis} then asks why this benefit erodes as budgets grow. Non-greedy revision of the acquired set did not recover accuracy, and accepted revisions were only weakly correlated with true predictive utility, indicating a mismatch between the diagnostic objective and the utility it approximates rather than greedy irreversibility alone. The picture that emerges is conditional, not uniform. Pairwise synergy can help, but only when the acquisition is both structurally present and budget-feasible. Improving a set-level score is not equivalent to improving downstream predictive performance.

We observe three distinct failure modes: unfinishable pairs with different costs under tight budgets (synthetic sweeps), state-independent acquisition when population-level scores dominate (MiniBooNE), and budget erosion with aggregated scores weakly aligned with predictive utility (PhysioNet, MNIST-loop). PhysioNet supports the existence of relevant pairwise structure in real data via permutation tests, but removing the pair branch does not change its acquisition gains essentially. The synthetic controls more directly connect information composition and feasibility to relative performance. Together, these findings suggest why synergistic/redundant pairs alone are insufficient. Deploying them also requires enough budget to complete the acquisition, an acquisition rule that responds to realized values rather than collapsing to population-level scores, and an aggregation objective that aligns with downstream utility as more features are acquired.

MiniBooNE shows a strong negative case. Its state-dependent ranking collapses to an effectively static acquisition pattern dominated by population-level scores rather than realized conditional values. SynAFA acquires the identical feature subset for all test instances at every tested budget, unlike the other four tabular datasets. Holding the acquired features fixed, retraining the classifier on the real masks substantially narrows the gap to AACO, CAE, and permutation, though most comparisons remain significant losses (Table~\ref{tab:miniboone-classifier-realignment}). However, these losses do not imply that loss of adaptivity causes the residual gap. Our acquisition comparisons and objective alignment results therefore concern utility under the evaluation classifier, which can favor some mask distributions over others. Appendix~\ref{app:miniboone} gives the trajectory, binarization, and sample-size diagnostics.

The evidence concerns the tested policy, aggregation, datasets, and budgets. The fixed-$V$ control covers the upper range of the synergy fraction. The post-hoc feasibility mask requires prospective validation. Our implementation binarizes features and assumes a binary outcome. Pairwise PID cannot represent interactions requiring three or more sources, while higher-order extensions introduce computational challenges. Amortized bias and variance are not fully characterized, and the reduced-dimensional exact control changes more than the amortized type. The real-data benchmarks use uniform costs, while heterogeneous costs are tested only in the controlled sweep. Bootstrap tests are not adjusted for multiple comparisons, and the small test sets of CKD and ACTG175 limit precision. Benchmark policies use a single training seed, so the bootstrap reflects test sampling, not training variability. The shared classifier supports comparisons under a fixed prediction model, rather than implying that the predictive utility would hold with a different classifier.

The results motivate evaluating whether candidate scores align with incremental utility given the acquired set before investing in more extensive search. Prospective comparisons could combine feasibility within the remaining budget and objectives that discount already acquired information, and test robustness across prediction models and policy-specific mask distributions. These are design directions, not validated improvements. Higher-order PID remains complementary. The duplicate-copy construction shows that this aggregation can fail even without higher-order synergy. Atom-specific reweighting also requires an active pair branch, so its effect depends on the weight of acquisition costs. Also, it does not generally rescue the negative benchmark pattern (Appendix~\ref{app:lambda-gating}).

\section{Conclusion}
We studied when pairwise information structure can support active feature acquisition through SynAFA, a policy combining joint information and realized-value conditional information. Across the evaluated benchmark datasets, its benefits are heterogeneous. Feature structure supported by permutation tests accompanies its strongest gains on PhysioNet at low budgets, which do not depend on the pair branch. However, these gains do not extend consistently across datasets or budgets. Controlled sweeps provide more targeted evidence. Within the tested synthetic family, SynAFA's advantage increases as pairwise information becomes more synergy-dominated, provided the relevant acquisition is feasible in terms of budget. Additionally, pair branch ablation shows that this benefit comes from pair proposals, which contribute little on most tested real datasets. A feasibility-mask diagnostic links the high-synergy losses under tight budgets to pairs that cannot be completed.

A distinct limitation emerges when local scores are aggregated into a set-level objective. Non-greedy single-swap local search improves the diagnostic objective, yet produces no accuracy gain in any of the tested settings. Accepted objective improvements are weakly correlated with changes in the predictive utility of the shared classifier. The duplicate-copy construction identifies one structural reason why this aggregation can overcount information. These findings are limited to the tested objective, single-swap local search, and the masked evaluation classifier. They do not imply that pairwise information or PID is ineffective more generally. They motivate acquisition objectives that jointly account for feasibility within the remaining budget, information already available in the acquired subset, and downstream predictive utility.
\clearpage

\clearpage
\bibliography{iclr2027_conference}

@article{williams2010nonnegative,
  title={Nonnegative decomposition of multivariate information},
  author={Williams, Paul L and Beer, Randall D},
  journal={arXiv preprint arXiv:1004.2515},
  year={2010}
}

@article{wollstadt2023rigorous,
  title={A rigorous information-theoretic definition of redundancy and relevancy in feature selection based on (partial) information decomposition},
  author={Wollstadt, Patricia and Schmitt, Sebastian and Wibral, Michael},
  journal={Journal of Machine Learning Research},
  volume={24},
  number={131},
  pages={1--44},
  year={2023}
}

@article{ma2018eddi,
  title={Eddi: Efficient dynamic discovery of high-value information with partial vae},
  author={Ma, Chao and Tschiatschek, Sebastian and Palla, Konstantina and Hern{\'a}ndez-Lobato, Jos{\'e} Miguel and Nowozin, Sebastian and Zhang, Cheng},
  journal={arXiv preprint arXiv:1809.11142},
  year={2018}
}

@inproceedings{zannone2019odin,
  title={Odin: Optimal discovery of high-value information using model-based deep reinforcement learning},
  author={Zannone, Sara and Hern{\'a}ndez-Lobato, Jos{\'e} Miguel and Zhang, Cheng and Palla, Konstantina},
  booktitle={ICML Real-world Sequential Decision Making Workshop},
  year={2019}
}

@article{shim2017pay,
  title={Why pay more when you can pay less: A joint learning framework for active feature acquisition and classification},
  author={Shim, Hajin and Hwang, Sung Ju and Yang, Eunho},
  journal={arXiv preprint arXiv:1709.05964},
  year={2017}
}

@inproceedings{balin2019concrete,
  title={Concrete autoencoders: Differentiable feature selection and reconstruction},
  author={Bal{\i}n, Muhammed Fatih and Abid, Abubakar and Zou, James},
  booktitle={International conference on machine learning},
  pages={444--453},
  year={2019},
  organization={PMLR}
}

@inproceedings{covert2023learning,
  title={Learning to maximize mutual information for dynamic feature selection},
  author={Covert, Ian Connick and Qiu, Wei and Lu, Mingyu and Kim, Na Yoon and White, Nathan J and Lee, Su-In},
  booktitle={International Conference on Machine Learning},
  pages={6424--6447},
  year={2023},
  organization={PMLR}
}

@inproceedings{gadgil2024estimating,
  title={Estimating conditional mutual information for dynamic feature selection},
  author={Gadgil, Soham and Covert, Ian and Lee, Su-In},
  booktitle={International Conference on Learning Representations},
  volume={2024},
  pages={34979--35010},
  year={2024}
}

@article{valancius2023acquisition,
  title={Acquisition conditioned oracle for nongreedy active feature acquisition},
  author={Valancius, Michael and Lennon, Max and Oliva, Junier},
  journal={arXiv preprint arXiv:2302.13960},
  year={2023}
}

@inproceedings{schutz2026afabench,
  title={Afabench: A generic framework for benchmarking active feature acquisition},
  author={Sch{\"u}tz, Valter and Wu, Han and Rezvan, Reza and Aronsson, Linus and Haghir Chehreghani, Morteza},
  booktitle={Proceedings of the 32nd ACM SIGKDD Conference on Knowledge Discovery and Data Mining V. 2},
  pages={9765--9776},
  year={2026}
}

@article{aronsson2025survey,
  title={A survey on active feature acquisition strategies},
  author={Aronsson, Linus and Rahbar, Arman and Chehreghani, Morteza Haghir},
  journal={arXiv preprint arXiv:2502.11067},
  year={2025}
}

@article{mnih2015human,
  title={Human-level control through deep reinforcement learning},
  author={Mnih, Volodymyr and Kavukcuoglu, Koray and Silver, David and Rusu, Andrei A and Veness, Joel and Bellemare, Marc G and Graves, Alex and Riedmiller, Martin and Fidjeland, Andreas K and Ostrovski, Georg and others},
  journal={nature},
  volume={518},
  number={7540},
  pages={529--533},
  year={2015},
  publisher={Nature Publishing Group UK London}
}

@inproceedings{silva2012predicting,
  title={Predicting in-hospital mortality of icu patients: The physionet/computing in cardiology challenge 2012},
  author={Silva, Ikaro and Moody, George and Scott, Daniel J and Celi, Leo A and Mark, Roger G},
  booktitle={2012 computing in cardiology},
  pages={245--248},
  year={2012},
  organization={IEEE}
}

@article{lecun1998gradient,
  title={Gradient-based learning applied to document recognition},
  author={LeCun, Yann and Bottou, L{\'e}on and Bengio, Yoshua and Haffner, Patrick},
  journal={Proceedings of the IEEE},
  volume={86},
  number={11},
  pages={2278--2324},
  year={1998},
  publisher={Ieee}
}

@article{huang2025information,
  title={Information Templates: A New Paradigm for Intelligent Active Feature Acquisition},
  author={Huang, Hung-Tien and Dinh, Dzung and Oliva, Junier B},
  journal={arXiv preprint arXiv:2508.18380},
  year={2025}
}
\bibliographystyle{iclr2027_conference}
\clearpage
\renewcommand{\thefigure}{A\arabic{figure}}
\setcounter{figure}{0}
\renewcommand{\thetable}{A\arabic{table}}
\setcounter{table}{0}
\appendix
\section{Appendix}

\subsection{Preprocessing, protocol, and baseline details}
\label{app:preprocessing-protocol}
All comparisons use a \emph{hard-budget-matched, paired bootstrap} significance test. Both methods are evaluated under the same hard acquisition budget, so cost is matched by construction rather than by nearest-cost approximation across differing $\lambda$ grids. All reported results use the corrected evaluation harness, which enforces this cost-matching for every paired comparison; an earlier harness issue that could leave it unenforced for methods without their own forced-acquisition logic, and its effect on the affected results, is documented in Appendix~\ref{app:full-tables}. For the v1/v2 datasets below, per-feature cost is uniform, so this also means both methods acquire exactly the same number of features; the controlled synergy-fraction sweep (\S\ref{sec:dose-response}) instead uses a deliberately heterogeneous-cost feature pair, where the same hard budget caps total cost rather than feature count. SynAFA's acquisition-cost weight $\lambda$ is 0.01 for all tabular and synthetic hard-budget experiments, without dataset-specific tuning and not selected using validation- or test-set performance (Appendix~\ref{app:lambda-gating} examines its role for the pairwise branch). For the image-scale MNIST-loop task (SynAFA v2), $\lambda=0.001$ was fixed before any MNIST-loop result was inspected, because pixel features are sparser than tabular features (median marginal information 0.0014 bits per pixel); a post hoc sensitivity analysis over $\lambda\in\{0.001,0.003,0.01\}$ (Appendix~\ref{app:lambda-mnist}) shows that 0.001 gives the highest SynAFA accuracy of the three values tested, and that the comparisons with AACO and CAE keep their sign at every $\lambda$. The two methods are evaluated with the \emph{same externally trained classifier}, which makes the final prediction from the acquired features, thereby controlling prediction-model identity across policies. This does not remove differences in classifier generalization across acquired-mask distributions (Appendix~\ref{app:miniboone}). The classifier is trained once per dataset with a random feature-masking curriculum so that this single fixed model can accept arbitrary observed subsets, rather than only the full feature vector; training on random masks does not guarantee equal accuracy across all policy-specific mask distributions (architecture and training details in Appendix~\ref{app:external-classifier}). For each test instance $i$, let $d_i = \mathbf{1}(\hat
y_i^{\mathrm{SynAFA}}=y_i) - \mathbf{1}(\hat
y_i^{\mathrm{baseline}}=y_i)$; the reported effect size is
$\Delta\mathrm{Acc} = \mathrm{Acc}_{\mathrm{SynAFA}} -
\mathrm{Acc}_{\mathrm{baseline}} = \frac{1}{n}\sum_i d_i$. Accuracy is measured with the class-weighted external classifier; on PhysioNet and BankMarketing, every evaluated method lies below the majority-class rate (Table~\ref{tab:dataset-overview}), so only paired differences between policies are interpreted. Performance differences are paired by test instance, and statistical significance is assessed using a two-sided paired bootstrap test (10{,}000 resamples of instance indices, applied jointly to both methods). Reported $p$-values are per-comparison tests, not adjusted for multiplicity across datasets, budgets, or baselines; we therefore interpret the repeated cross-budget/baseline patterns rather than isolated significance calls.

\begin{figure}[h]
\centering
\includegraphics[width=0.85\linewidth]{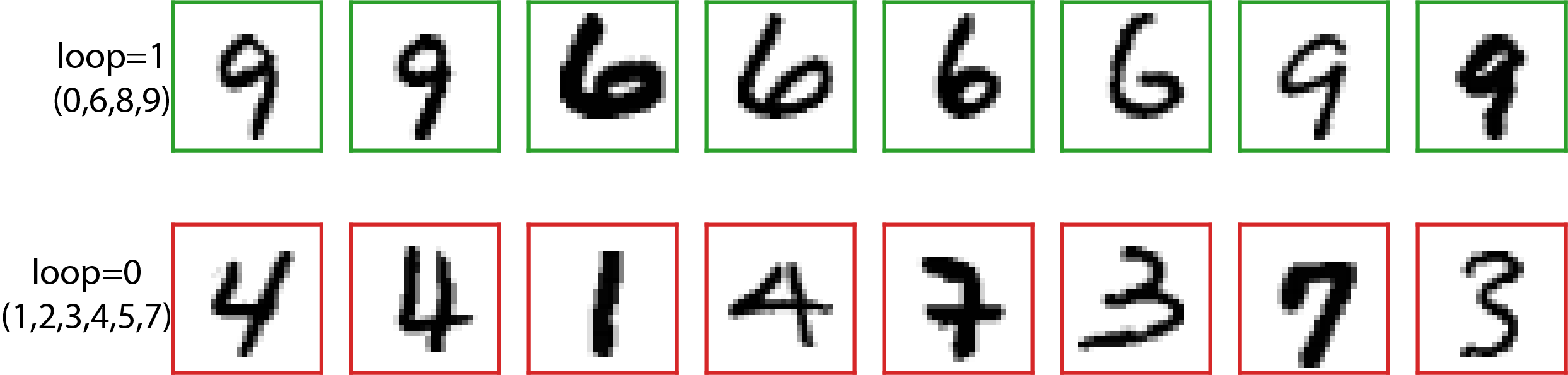}
\caption{Real examples from the MNIST-loop binary relabeling used for $D=784$ experiments.}
\label{fig:mnist-loop}
\end{figure}

\textbf{Datasets (v1, tabular):} CKD, ACTG175, BankMarketing, PhysioNet
\citep{silva2012predicting}, and MiniBooNE. \textbf{Dataset (v2, image-scale):} MNIST \citep{lecun1998gradient}, relabeled as a binary ``closed loop'' task (digits $\{0,6,8,9\}$ vs.\ $\{1,2,3,4,5,7\}$; Fig.~\ref{fig:mnist-loop}). This task was chosen because identifying a closed loop requires spatially-separated pixel regions to be jointly ``on'', a visually interpretable setting in which spatially separated pixels can carry joint predictive information; a parity relabeling, by contrast, has no such visual correlate. Appendix~\ref{app:datasets} (Table~\ref{tab:dataset-overview}) summarizes each dataset's prediction task, domain, dimensionality, and test set size.

\textbf{Baselines:} AACO \citep{valancius2023acquisition}, a permutation-importance baseline, CAE \citep{balin2019concrete}, GDFS \citep{covert2023learning}, DIME \citep{gadgil2024estimating}, JAFA \citep{shim2017pay}, and OL (a model-free DQN-based non-greedy RL baseline, with and without an observation-mask input) on all five v1 datasets; EDDI \citep{ma2018eddi} and ODIN \citep{zannone2019odin} on the two smallest datasets only. Under our evaluation configuration, their partial-VAE-based acquisition scoring uses 128 Monte Carlo forward passes per candidate feature per step, making exhaustive evaluation on the three larger datasets computationally prohibitive.

\paragraph{Splits, binarization and decision rules (read from the implementation).}
All datasets, real and synthetic, use the default 60/20/20 train/validation/test split (the test-set sizes in Table~\ref{tab:dataset-overview} are the remaining 20\%, e.g., 2{,}400 of 12{,}000 PhysioNet patients). Dataset instances $0,\dots,4$ are generated and split with seeds $42,\dots,46$; methods are trained with seed 0 on instance 0 (seed $i$ on synthetic instance $i$). SynAFA binarizes each feature with a threshold fitted on the training data: the midpoint of the two values for a two-valued column, a constant for a single-valued column, and otherwise the median, falling back to the largest value strictly below the maximum whenever the median split would be degenerate (all values on one side); a value maps to 1 iff it exceeds its threshold. Ties are broken toward the lowest feature index: single-feature candidates are compared in ascending index order with a strict inequality, a pair proposal replaces the best single-feature proposal only if its score is strictly larger, and when a pair wins, the member with the larger marginal information is acquired (the lower-index member on ties). If the policy proposes to stop before the hard budget is exhausted, the harness acquires the lowest-index unobserved feature.

\subsection{PID definitions}
\label{app:pid-definitions}
For a discrete source $X$ and target value $y$, using base-two logarithms,
\[
I_{\rm spec}(y;X)=\sum_x p(x\mid y)\log_2\frac{p(y\mid x)}{p(y)}.
\]
The Williams--Beer redundancy and remaining bivariate atoms are
\begin{align*}
R&=\sum_y p(y)\min\{I_{\rm spec}(y;X_j),I_{\rm spec}(y;X_k)\},\\
U_j&=I(X_j;Y)-R,\qquad U_k=I(X_k;Y)-R,\\
\mathrm{Syn}&=I(X_j,X_k;Y)-I(X_j;Y)-I(X_k;Y)+R.
\end{align*}
These recover Eq.~\ref{eq:pid-main}. The deployed pair score is their sum $V_{jk}$; atom reweighting is a separate diagnostic (Appendix~\ref{app:lambda-gating}). Features use median-split binarization with a safeguard for near-binary or highly skewed columns, as in the source implementation.

\subsection{PhysioNet validation and PID illustration}
\label{app:physionet-validation}
\begin{figure}[h]
\centering
\includegraphics[width=0.85\linewidth]{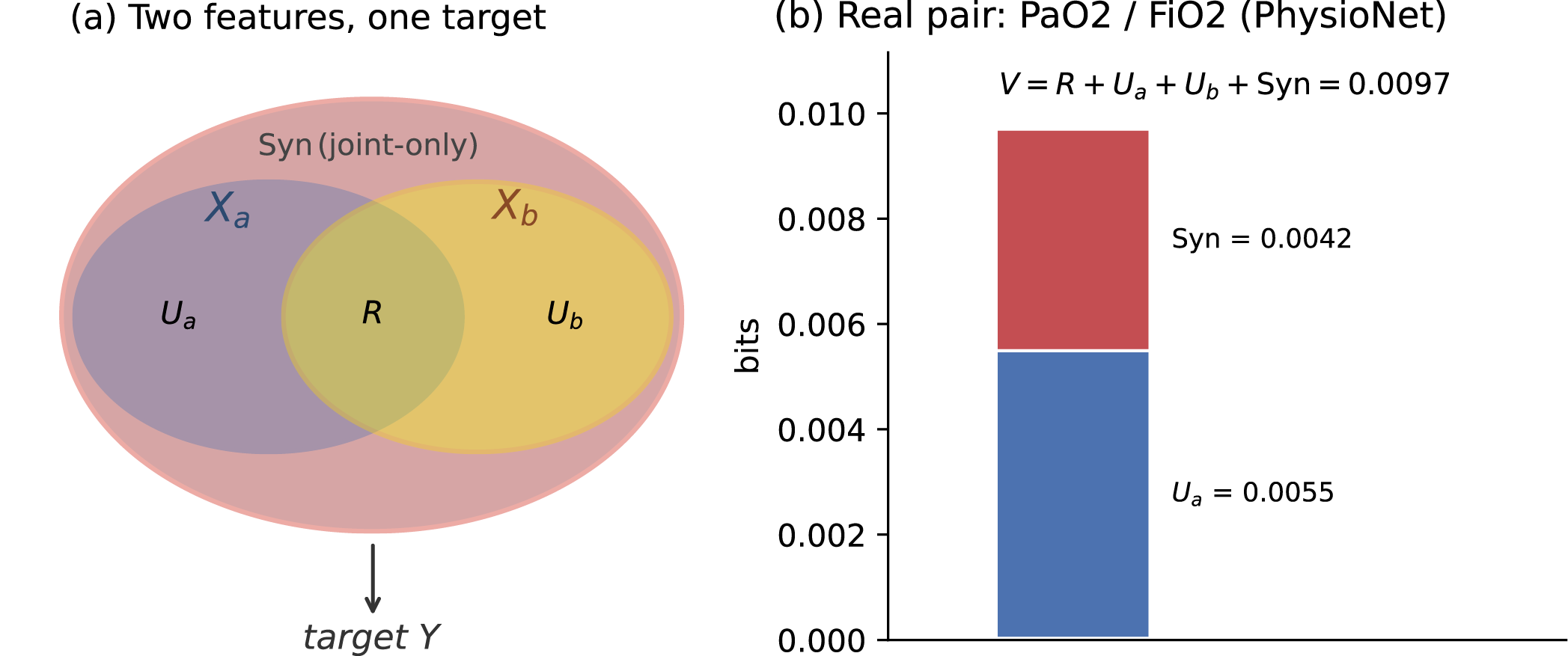}
\caption{(a) Schematic: two features $X_a, X_b$ jointly informing target $Y$, decomposed into redundant ($R$), unique ($U_a$, $U_b$), and
synergistic ($\mathrm{Syn}$) information. (b) The decomposition applied to a real PhysioNet pair, PaO$_2$/FiO$_2$: this particular pair happens to have $R \approx U_b \approx 0$ (PaO$_2$ dominates FiO$_2$'s unique contribution almost entirely), with a non-zero synergistic component in addition to the dominant unique contribution from PaO$_2$.}
\label{fig:pid-schematic}
\end{figure}

Before evaluating the AFA policies, we tested for synergistic and redundant feature structure in PhysioNet using a
stratified permutation test, with permutations performed within outcome strata to preserve class-specific marginal distributions. We tested this in two ways. First, we scanned all usable feature pairs (378 pairs among the 28 variables with under 35\% missingness; the 351 with at least 200 jointly observed patients were tested) on the 12{,}000-patient cohort, each pair on its complete cases (n$=8{,}970$ for GCS/PaO$_2$; the cohort includes the patients later used to evaluate the policies, and SynAFA never uses the scan output), with Benjamini--Hochberg FDR correction at $q=0.05$ (500 within-stratum permutations per pair for screening, 2{,}000 for pairs passing the screen); GCS (level of consciousness) and PaO$_2$ (oxygenation) was the strongest synergy pair ($\mathrm{Syn}=0.0127$ bits, raw permutation $p=0.0005$, the smallest value attainable with 2{,}000 permutations). Second, as a held-out robustness check, we repeated this scan on a randomly split discovery half alone (6{,}000 patients per half; 51 of 351 pairs surviving FDR there), then tested only those discovery-surviving pairs on the completely held-out validation half, with FDR correction applied across that confirmatory set; 36 of these 51 pairs (70.6\%) replicated, including GCS/PaO$_2$. BUN and creatinine, a clinically motivated candidate pair tested separately from this scan, provide a confirmed example of redundant information (the pair was specified from clinical reasoning before any scan was run, whereas GCS/PaO$_2$ emerged from the scan and was not pre-specified). To illustrate how these different information components are represented within the PID framework, Figure~\ref{fig:pid-schematic} shows the decomposition of another real PhysioNet feature pair, PaO$_2$ and FiO$_2$, whose joint information contains a visible synergistic component.

\subsection{Controlled-sweep construction and secondary results}
\label{app:synthetic-details}
The PhysioNet result provides real-data evidence but does not isolate the role of pairwise information structure from other dataset characteristics. We therefore constructed a controlled synthetic setting in which the information structure of one designated feature pair is varied systematically while the remaining data-generating process is held fixed. Concretely, we built a synthetic, $D=24$, binary-outcome dataset with that designated feature pair, S1 (cost 1) and S2 (cost 5, motivated by cheap versus expensive clinical measurements; the real-data benchmark itself uses uniform costs), and swept a parameter $\alpha \in \{0, 0.25, 0.5, 0.75, 1\}$ controlling only the pair's \emph{interaction structure}: at $\alpha=0$ the pair's contribution to the outcome is purely additive (S1 and S2 each independently shift the outcome's log-odds, so each carries real marginal information on its own); at $\alpha=1$ it is pure XOR synergy (neither S1 nor S2 shifts the outcome's marginal distribution at all, only their joint value together does). Concretely, the pair's contribution to the outcome logit is
\[
(1-\alpha)\big[w_1(2S_1{-}1) + w_2(2S_2{-}1)\big] + \alpha\, w_{\mathrm{xor}}\big[2(S_1 \oplus S_2){-}1\big],
\]
with $w_1=w_2=0.7$ and $w_{\mathrm{xor}}=1.4$ held fixed across $\alpha$, along with the other 22 features (independent, decreasing individual informativeness); only the interpolation weight $\alpha$ between the two bracketed terms changes, i.e., the interaction structure, not either term's own coefficient scale (the resulting joint information $V$ is consequently not itself monotonic in $\alpha$, only its synergistic fraction is; see Table~\ref{tab:synergy-pid-decomposition}). Each of the five $\alpha$ settings was run through the identical official pipeline used for every other result in this paper: SynAFA and three independently implemented, independently trained baselines (CAE, AACO, permutation) at hard budgets of 3, 5, and 8, an external classifier, $n=6{,}000$ test instances per comparison and dataset instance (five independently generated dataset instances per $\alpha$; Appendix~\ref{app:synthetic-generative}).

Table~\ref{tab:synergy-dose-response} and Figure~\ref{fig:synergy-dose-response} report the result. The direction of the effect depends sharply on whether the designated pair is affordable. At budget 8, SynAFA's mean advantage increases as the pair becomes increasingly synergy-dominated, from a near-zero mean difference at $\alpha=0$ ($+0.007$, five-instance mean) to a substantial positive mean advantage at $\alpha=1$ ($+0.077$, instance-bootstrap 95\% interval $[0.063,0.089]$; up to $+0.140$ against CAE individually, with permutation at parity), whereas at budgets 3 and 5 the trend reverses and SynAFA's relative performance deteriorates, with significant losses against individual baselines at high $\alpha$. The corresponding descriptive correlations across the five prescribed $\alpha$ levels are $+0.946$, $-0.974$, and $-0.945$, respectively (budgets 8, 3, 5), and each has the same sign in all five dataset instances (per-instance ranges $[+0.79,+0.97]$, $[-0.99,-0.77]$, $[-0.99,-0.87]$).

The acquisition traces reveal the mechanism behind this reversal (Table~\ref{tab:synergy-acquisition-rates}). When the pair is unaffordable, increasing $\alpha$ shifts SynAFA away from useful standalone acquisitions: at budget 3 it abandons the inexpensive S1 (in four of five dataset instances at $\alpha=1$), while at budget 5 it acquires the expensive S2 alone at high $\alpha$ (four of five instances at $\alpha=1$). At budget 8, where the full pair is feasible, both features are acquired in 100\% of instances, in every dataset instance, at $\alpha\ge0.75$ (at $\alpha=0$ completion is instance-dependent, 0--100\%, mean 30\%), alongside the improvement in relative accuracy.

\textbf{Is this a general synergy-feasibility interaction, or an artifact of the deployed pair-proposal rule?} SynAFA's pair branch (\S\ref{sec:method}) scores an unstarted pair by its cost-adjusted value without checking whether the remaining budget can actually afford to complete it, so it can propose, and acquire one member of, a pair it can never finish. To test whether the negative trend at budgets 3 and 5 reflects this specific mechanism gap rather than a general property of synergy under tight budgets, we reran the identical five-$\alpha$ sweep with one targeted change: the pair branch is masked to only ever propose a pair the remaining budget can afford, tested here only and nowhere else in this paper. The masked variant is evaluated with a re-implementation of the harness's hard-budget loop, which reproduces the official deployed results exactly in all 75 instance-by-$\alpha$-by-budget cells. Masking changes budget 8 in only 3 of the 25 instance-by-$\alpha$ cells (largest change in mean advantage 0.009); Table~\ref{tab:synergy-dose-response} is otherwise unchanged. At budgets 3 and 5, it removes the high-$\alpha$ negative mean effects (Table~\ref{tab:synergy-budget-mask-sensitivity}): mean advantage is positive at every tested $\alpha$, without being monotonic (e.g., budget 3, five-instance means $+0.014$ $\to$ $+0.048$ at $\alpha=0.5$, $-0.045$ $\to$ $+0.065$ at $\alpha=0.75$, $-0.051$ $\to$ $+0.064$ at $\alpha=1$) instead of degrading into significant losses, and the same qualitative removal of the adverse high-$\alpha$ pattern is visible in the per-baseline comparisons (Appendix~\ref{app:full-tables}, Table~\ref{tab:synergy-budget-mask-sensitivity-full}). The recovery from high-$\alpha$ losses under the masked rule supports pair-level budget infeasibility as an explanation for those losses, rather than establishing a general adverse effect of synergy under tight budgets. We retain the deployed rule for all primary benchmark comparisons to avoid redefining the method after observing the synthetic failure mode; the masked variant is used only to identify whether pair-level budget infeasibility is responsible for the observed reversal. Pair-level feasibility masking is therefore best viewed as a post hoc mechanism diagnostic rather than a validated algorithmic improvement; evaluating it prospectively as a candidate method change is left for future work.

Table~\ref{tab:synergy-pid-decomposition} confirms that the manipulation changes the composition, rather than simply the magnitude, of the pair's joint information. Under the symmetric construction, $U_a=U_b=0$ for the Williams--Beer $I_{\min}$ decomposition. Neither $\mathrm{Syn}$ nor $V$ increases monotonically with $\alpha$; instead, the synergy fraction $\mathrm{Syn}/V$ rises monotonically from 0.52 to 1.00. We therefore interpret $\alpha$ as controlling how synergy-dominated the pairwise information becomes, rather than as directly controlling total joint information.

Together, these results show that the positive trend at budget 8, where the pair is affordable in isolation, is the consistent finding within this controlled setting: SynAFA's advantage rises with the pair's synergy fraction at this tested budget, and this result is essentially unchanged by the feasibility mask (Table~\ref{tab:synergy-dose-response}). At budgets 3 and 5, masking diagnoses a contribution from pair-level budget infeasibility to high-synergy losses. Positive masked effects need not have a positive trend across the sweep, so this control should not be read as establishing a general monotonic benefit under tight budgets. Because increasing $\alpha$ simultaneously shifts information from standalone to joint form, this synthetic manipulation characterizes the consequences of increasingly synergy-dominated pairwise information structure, rather than isolating the causal effect of the PID synergy atom while holding all other information components fixed. Full per-baseline results for both the deployed and masked rules are in Appendix~\ref{app:full-tables} (Tables~\ref{tab:synergy-dose-response-full} and \ref{tab:synergy-budget-mask-sensitivity-full}); the exact PID decomposition and acquisition-rate breakdown are in Appendix~\ref{app:full-tables} (Tables~\ref{tab:synergy-pid-decomposition} and \ref{tab:synergy-acquisition-rates}).

\textbf{A fixed-$V$ synergy-fraction control.} This leaves open whether the budget-8 trend persists when the pair becomes more synergy-dominated without increasing its total joint information $V$, or simply reflects that $V$ happens to be larger at high $\alpha$. To separate these, we built a second, calibrated sweep: for five target synergy fractions $\mathrm{Syn}/V \in \{0.52, 0.65, 0.78, 0.90, 1.00\}$, the additive and XOR weights ($w_1{=}w_2$ and $w_{\mathrm{xor}}$ above) were numerically recalibrated to hold $V$ approximately constant while $\mathrm{Syn}/V$ varies, reusing the same symmetric generative family as the original sweep. Calibration was verified on the actual generated data, not only at the population level: realized $V$ ranged from 0.1003 to 0.1032 bits (a spread under 3\%) while realized $\mathrm{Syn}/V$ ranged from 0.512 to 1.000 (Table~\ref{tab:fixed-v-calibration}, Appendix~\ref{app:full-tables}). Numerical small-effect checks in the symmetric family approach $\mathrm{Syn}/V=0.5$ (Table~\ref{tab:fixed-v-calibration}), so this control covers the upper half of the synergy-fraction range rather than $[0,1]$ in full.

At budget 8, SynAFA's mean advantage over AACO, CAE, and permutation rose from $+0.010$ at $\mathrm{Syn}/V=0.51$ to $+0.028$ at $\mathrm{Syn}/V=1.00$ (Pearson $r=0.874$ between realized synergy fraction and mean advantage; Figure~\ref{fig:synergy-dose-response}c), with the trend strongest against AACO and CAE (rising to $+0.043$ and $+0.041$, respectively) and the permutation baseline remaining near parity throughout, consistent with the original sweep (full per-baseline results and significance in Table~\ref{tab:fixed-v-full-significance}, Appendix~\ref{app:full-tables}). Because $V$ is held nearly constant across this sweep, the increasing advantage cannot be explained by the pair simply carrying more total joint information. SynAFA's advantage increases as joint information becomes more synergy-dominated, including when its total magnitude is held approximately constant.

\subsection{MNIST-loop estimator and policy controls}
\label{app:mnist-diagnostics}
We tested four candidate explanations for the CAE performance gap, using a direct diagnostic for each rather than relying on qualitative interpretation.

\textbf{The hard-budget fallback does not explain the gap.}
We first tested whether the hard-budget evaluation itself could explain the observed performance gap. Under this protocol, if a policy chooses to stop before exhausting its budget, an additional feature is selected by a fallback rule. However, across all 12{,}000 test instances at each of the three budgets, SynAFA v2 never stopped early: its acquisition scores remained above the stopping threshold until the budget was exhausted. The hard-budget fallback was therefore never invoked and cannot explain the observed gap.

\textbf{State-dependent acquisition, not static pairwise scoring, accounts for the gain over marginal selection.}
We constructed two non-adaptive ablations from the same trained (sampling-corrected) table, evaluated on the full test set: static-marginal selection using \texttt{single\_value} alone, and
static-pairwise selection that retains pairwise scoring but does not adapt acquisition scores to the realized values of previously acquired features. All policies were evaluated with the same external classifier. Accuracy on the full test set at budgets 5, 10 and 20 was 0.671, 0.744 and 0.766 for static-marginal selection, 0.708, 0.726 and 0.756 for static-pairwise selection, and 0.748, 0.776 and 0.805 for the full adaptive policy. Realized-value adaptation accounts for the improvement over both static selections at every budget, whereas static pairwise scoring alone improves on marginal ranking only at budget 5 and not at budgets 10 and 20.

\textbf{Initial local-search diagnostic (separate from the nine-setting test).}
We next allowed the acquired set to revise earlier selections through a corrected local search that exactly recomputes the full set-level objective for every candidate swap, guaranteeing monotonic improvement in the objective. The search converged for all instances within 200 rounds. Surprisingly, it increased the objective from 2.60 to 3.77 while accuracy decreased from 0.836 to 0.806 (Table~\ref{tab:local-search-progression}). Thus, stronger optimization of this diagnostic set-level aggregation did not resolve the MNIST-loop performance gap, motivating us to examine whether the aggregation itself is aligned with predictive utility.

\begin{figure}[h]
\centering
\includegraphics[width=0.85\linewidth]{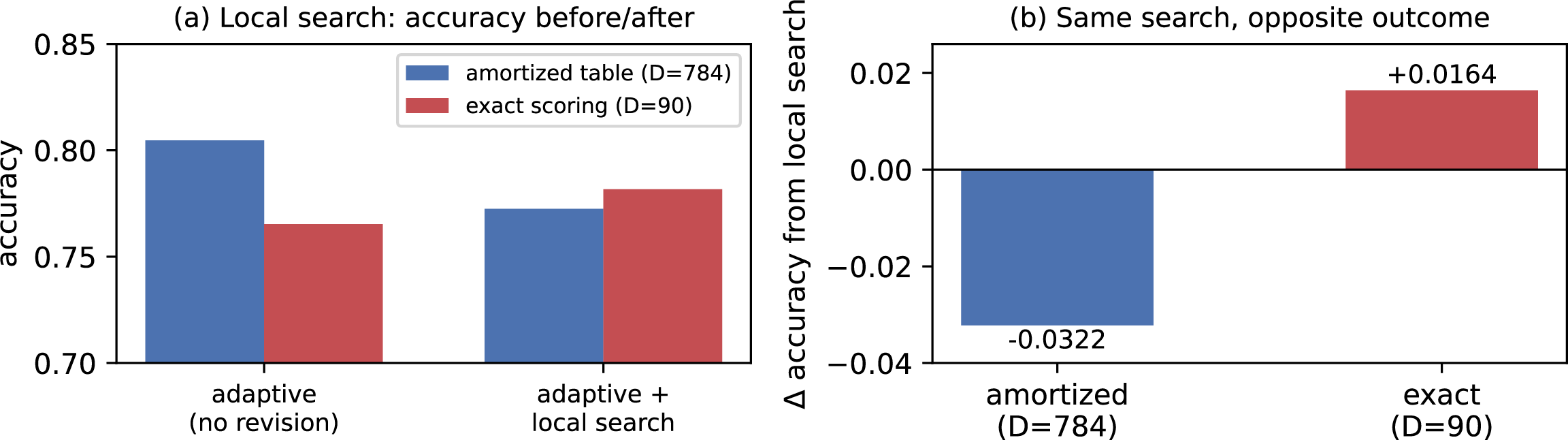}
\caption{A diagnostic test for whether amortized estimation error contributes to the reversal in this specific comparison (\S\ref{sec:analysis}): the same single-swap local-search procedure, run against an amortized table ($D=784$, left/blue) vs.\ an exact-information table ($D=90$, right/red). Both conditions use the full 12{,}000-instance test set at budget 20. Dimensionality and estimator type change together between these two conditions, so this is diagnostic evidence for estimation error's role, not a controlled isolation of it. (a) Accuracy before/after search. (b) In this comparison the search reverses direction depending on whether the underlying scores are exact or learned (budget 20 shown; with exact scoring the search also improves accuracy at budget 5, $+0.0767$, but reduces it at budget 10, $-0.0125$), though \S\ref{sec:objective-alignment}'s broader, nine-setting test finds that changes in the pairwise aggregation objective are weakly aligned with changes in predictive utility even under exact scoring, so this $D=90$ result should not be interpreted as evidence that the scoring framework is generally effective.}
\label{fig:si-exact-vs-amortized}
\end{figure}

\textbf{Amortized estimation error is a contributing factor.}
On a reduced $D=90$ pixel subset, the accuracy gap between amortized and exact scoring nearly disappears as the precompute sample size increases ($0.0128\to0.0005$; Table~\ref{tab:amortized-k-infer}), and the same corrected local search that fails on the amortized $D=784$ table instead improves accuracy under exact $D=90$ scoring at budgets 5 and 20 ($\Delta\mathrm{Acc}=+0.0767$ and $+0.0164$, both $p<0.001$; full test set), though not at budget 10 ($-0.0125$; Fig.~\ref{fig:si-exact-vs-amortized}). Estimation error is therefore a plausible, though not isolated, contributor to the $D=784$ gap: this comparison changes dimensionality and estimator type together, so neither result establishes that the underlying pairwise aggregation objective is generally aligned with predictive utility. Section~\ref{sec:objective-alignment} tests that question directly across datasets and budgets.

Increasing the precompute sample and correcting the full-sample procedure together changed the budget-5 CAE gap from $-0.077$ to $-0.023$. At fixed \texttt{k\_infer}$=36{,}000$, Table~\ref{tab:d784-full} reports budget-5 accuracy of 0.719 with replacement and 0.748 after correction. These contrasts should not be attributed to the sampling correction alone (Appendix~\ref{app:amortized}).

\begin{figure}[h]
\centering
\includegraphics[width=\linewidth]{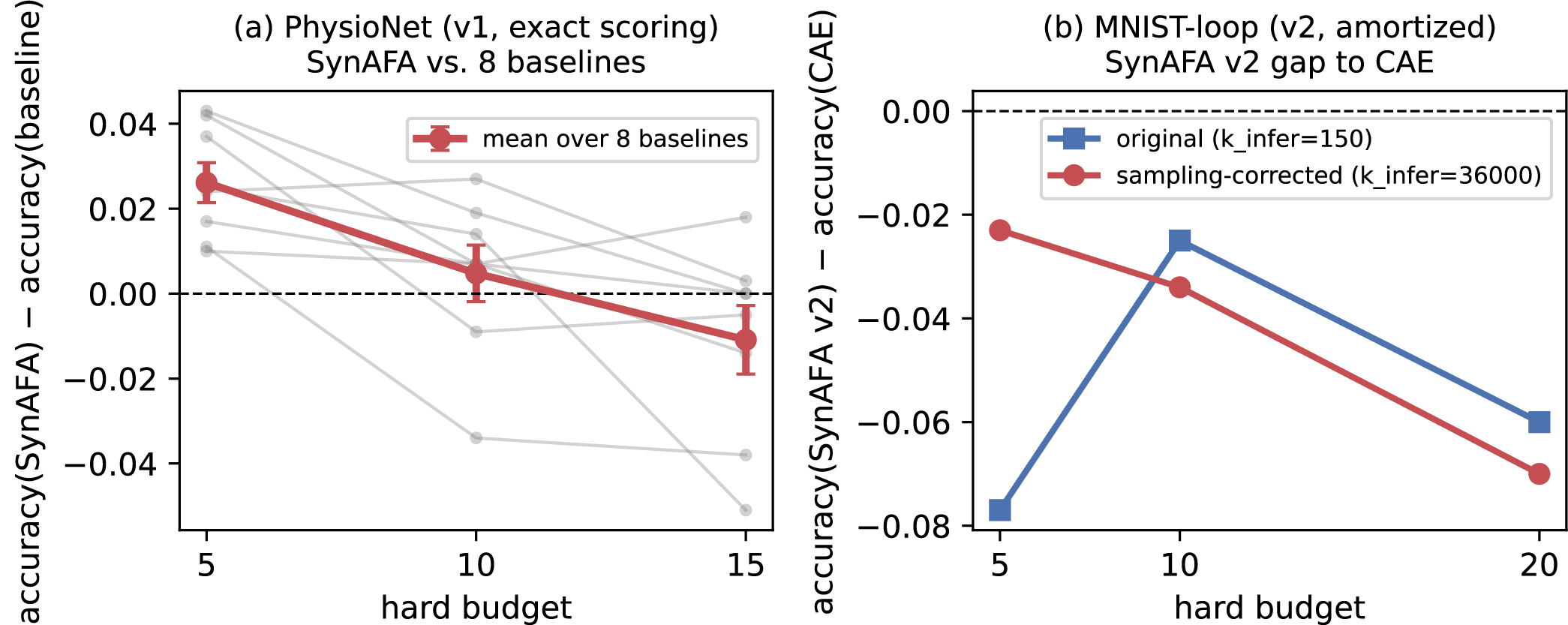}
\caption{A recurring budget-dependent erosion pattern across two
benchmark settings (\S\ref{sec:pattern}): SynAFA's relative performance
is strongest at low budgets and diminishes as budget grows. Greedy, one-step-ahead construction is a
natural candidate explanation for this pattern, tested directly in
\S\ref{sec:local-search-test}. (a) PhysioNet, SynAFA vs.\
each of 8 baselines (gray) and the mean $\pm$ SEM (red). (b)
MNIST-loop, SynAFA v2's accuracy gap to CAE before and after the
precomputation sampling correction (\S\ref{sec:analysis}).}
\label{fig:budget-erosion}
\end{figure}

\subsection{Dataset details}
\label{app:datasets}
Table~\ref{tab:dataset-overview} summarizes the prediction task, domain, feature dimensionality, and test set size for each of the six real-world datasets used in this paper.

\begin{table}[htbp]
\centering
\caption{Datasets and hard-budget test-set sizes. CKD, ACTG175, BankMarketing and MiniBooNE use the UCI dataset variants described by the benchmark. MNIST-loop is a binary relabeling. The purpose-built synthetic dataset is described in \S\ref{sec:dose-response}. The last column is the share of the more frequent class in the test set. The external classifier is trained with class-weighted cross-entropy (Appendix~\ref{app:external-classifier}) and predicts from partially observed features, so reported accuracies can lie below this rate: they do for every evaluated method at every tested hard budget on PhysioNet and BankMarketing, which is why only paired differences between policies are interpreted.}
\label{tab:dataset-overview}
\footnotesize
\setlength{\tabcolsep}{3pt}
\begin{tabular}{llp{4.4cm}rrr}
\multicolumn{1}{c}{\bf DATASET} & \multicolumn{1}{c}{\bf DOMAIN} & \multicolumn{1}{c}{\bf TASK (TARGET $Y$)} & \multicolumn{1}{c}{\bf $D$} & \multicolumn{1}{c}{\bf $n$} & \multicolumn{1}{c}{\bf MAJORITY} \\
\hline \\
CKD & clinical (nephrology) & chronic kidney disease present vs.\ absent & 24 & 80 & 0.600 \\
ACTG175 & clinical (HIV/AIDS trial) & treatment-failure / disease-progression indicator & 23 & 429 & 0.730 \\
BankMarketing & marketing / finance & client subscribed to a term deposit vs.\ not, following a phone campaign & 16 & 9{,}043 & 0.884 \\
PhysioNet & clinical (ICU / critical care) & in-hospital mortality of ICU patients \citep{silva2012predicting} & 41 & 2{,}400 & 0.862 \\
MiniBooNE & particle physics & electron-neutrino signal vs.\ muon-neutrino background event & 50 & 26{,}014 & 0.717 \\
MNIST-loop & vision (relabeled MNIST) & digit contains a topological loop (0,6,8,9) vs.\ not (1,2,3,4,5,7) \citep{lecun1998gradient} & 784 & 12{,}000 & 0.604 \\
\end{tabular}
\end{table}

\subsection{Reported benchmark comparisons and diagnostic tables}
\label{app:full-tables}

\textbf{Hard-budget evaluation harness correction.} The evaluation harness enforces cost-matching under a hard budget via a fallback rule: for methods that do not implement their own forced-acquisition logic (SynAFA among them), a stop decision issued before the target budget is exhausted should be overridden with the next available acquisition, so that voluntary early stopping cannot leave a method short of the budget it is nominally being compared at. We found that this fallback override was, due to an implementation error, never actually applied, letting SynAFA silently stop short of the target budget whenever its own policy chose to stop early. We corrected this and reran every affected hard-budget evaluation, confirming 100\% of test instances now reach the target budget wherever the method's own policy has no independent reason to fall short (\S\ref{sec:dose-response} documents one such reason: pair-level budget infeasibility in the controlled synergy-fraction sweep setting, which the correction does not and should not remove). The correction materially changed PhysioNet at budgets 10 and 15 (Table~\ref{tab:physionet-clean}: 80.25\% and 22.6\% of instances, respectively, previously reached the target budget, versus 100\% now), BankMarketing at budget 7 (17.7\% of instances previously reached the target budget), ACTG175 at budget 10 (90.2\%), and nine of the 25 SynAFA runs of the fixed-$V$ control (Table~\ref{tab:fixed-v-full-significance}). Among the reported benchmark comparisons, the corrected results preserve the qualitative conclusions, with effect-size shifts of at most 0.011; two PhysioNet comparisons lost significance (budget 10 against GDFS, budget 15 against permutation), and the budget-15 comparison with JAFA moved to the boundary of significance ($p\approx0.05$). In the fixed-$V$ control, the correction lowered the correlation between synergy fraction and mean advantage from $r=0.920$ to $r=0.874$ without changing its direction.

\begin{table}[htbp]
\centering
\caption{SynAFA versus AACO and CAE: accuracy differences under the shared classifier and matched hard budgets. $^*p<0.05$; unmarked values are not significant. Additional reported comparisons follow below. CKD ($n=80$) stars are bootstrap-based and are not significant under an exact sign test (Table~\ref{tab:ckd-actg}).}
\label{tab:v1-headline}
\footnotesize
\setlength{\tabcolsep}{4pt}
\begin{tabular}{lccc}
\multicolumn{1}{c}{\bf DATASET} &
\multicolumn{1}{c}{\bf BUDGET} &
\multicolumn{1}{c}{\bf VS.\ AACO} &
\multicolumn{1}{c}{\bf VS.\ CAE}
\\ \hline \\
CKD           & 2  & $-0.050^{*}$    & $+0.038$ (n.s.) \\
CKD           & 4  & $+0.000$ (n.s.) & $+0.075^{*}$ \\
CKD           & 7  & $+0.000$ (n.s.) & $+0.025$ (n.s.) \\
ACTG175       & 3  & $-0.023$ (n.s.) & $+0.016$ (n.s.) \\
ACTG175       & 5  & $-0.021$ (n.s.) & $-0.005$ (n.s.) \\
ACTG175       & 10 & $-0.009$ (n.s.) & $+0.009$ (n.s.) \\
BankMarketing & 2  & $+0.034^{*}$    & $+0.000$ (n.s.) \\
BankMarketing & 4  & $-0.017^{*}$    & $+0.015^{*}$ \\
BankMarketing & 7  & $-0.068^{*}$    & $-0.045^{*}$ \\
PhysioNet     & 5  & $+0.011$ (n.s.) & $+0.037^{*}$ \\
PhysioNet     & 10 & $-0.034^{*}$    & $-0.009$ (n.s.) \\
PhysioNet     & 15 & $-0.038^{*}$    & $-0.005$ (n.s.) \\
MiniBooNE     & 5  & $-0.034^{*}$    & $-0.025^{*}$ \\
MiniBooNE     & 10 & $-0.053^{*}$    & $-0.051^{*}$ \\
MiniBooNE     & 15 & $-0.033^{*}$    & $-0.044^{*}$ \\
\end{tabular}
\end{table}

\begin{table}[htbp]
\centering
\caption{PhysioNet accuracy(SynAFA)$-$accuracy(baseline), $n=2{,}400$, matched hard budgets and the shared classifier. $^*p<0.05$, unadjusted.}
\label{tab:physionet-clean}
\footnotesize
\setlength{\tabcolsep}{4pt}
\begin{tabular}{lccc}
\multicolumn{1}{c}{\bf BASELINE} &
\multicolumn{1}{c}{\bf BUDGET 5} &
\multicolumn{1}{c}{\bf BUDGET 10} &
\multicolumn{1}{c}{\bf BUDGET 15}
\\ \hline \\
AACO              & $+0.011$      & $-0.034^{*}$          & $-0.038^{*}$ \\
Permutation       & $+0.042^{*}$  & $+0.007$              & $-0.014$ \\
CAE               & $+0.037^{*}$  & $-0.009$              & $-0.005$ \\
GDFS              & $+0.025^{*}$  & $+0.014$              & $-0.051^{*}$ \\
DIME              & $+0.017^{*}$  & $+0.007$              & $-0.000$ \\
JAFA              & $+0.010$      & $+0.007$              & $+0.018^{*}$ \\
OL (with mask)    & $+0.024^{*}$  & $+0.027^{*}$          & $+0.003$ \\
OL (without mask) & $+0.043^{*}$  & $+0.019^{*}$          & $-0.000$ \\
\hline
{\bf Mean}        & ${\bf +0.026}$ & ${\bf +0.005}$        & ${\bf -0.011}$ \\
\end{tabular}
\end{table}

\begin{table}[htbp]
\centering
\caption{Deployed synthetic sweep: mean accuracy advantage over CAE, AACO and permutation, averaged over five dataset instances (standard deviation across instances in parentheses; $D=24$, $n=6{,}000$ test instances per instance). The designated pair costs 6, exceeding budgets 3 and 5. At budget 8, it is affordable in isolation; this does not guarantee completion along a trajectory. The last row correlates $\alpha$ with the five-instance mean; per-instance correlations range over budget 3: $[-0.99,-0.77]$; budget 5: $[-0.99,-0.87]$; budget 8: $[+0.79,+0.97]$. Full effects appear in Table~\ref{tab:synergy-dose-response-full}.}
\label{tab:synergy-dose-response}
\footnotesize
\setlength{\tabcolsep}{4pt}
\begin{tabular}{lccc}
\multicolumn{1}{c}{\bf $\alpha$} &
\multicolumn{1}{c}{\bf BUDGET 3} &
\multicolumn{1}{c}{\bf BUDGET 5} &
\multicolumn{1}{c}{\bf BUDGET 8}
\\ \hline \\
0.00 & $+0.060$ (0.036) & $+0.027$ (0.011) & $+0.007$ (0.006) \\
0.25 & $+0.038$ (0.024) & $+0.010$ (0.010) & $+0.030$ (0.006) \\
0.50 & $+0.014$ (0.017) & $-0.000$ (0.018) & $+0.028$ (0.012) \\
0.75 & $-0.045$ (0.020) & $-0.059$ (0.011) & $+0.045$ (0.006) \\
1.00 & $-0.051$ (0.022) & $-0.057$ (0.011) & $+0.077$ (0.016) \\
\hline
$\mathrm{corr}(\alpha,\ \mathrm{mean}\ \Delta)$ & $-0.974$ & $-0.945$ & $+0.946$ \\
\end{tabular}
\end{table}

\begin{table}[htbp]
\centering
\caption{Feasibility diagnostic: mean advantage (five dataset instances) under deployed and masked pair proposals. Masking recovers positive mean effects at high $\alpha$; it does not make advantage monotonic in $\alpha$. Budget 8 is essentially unchanged (identical in 22 of 25 instance-by-$\alpha$ cells) and is omitted. This variant is used only for the mechanism diagnostic.}
\label{tab:synergy-budget-mask-sensitivity}
\footnotesize
\setlength{\tabcolsep}{4pt}
\begin{tabular}{lcccc}
& \multicolumn{2}{c}{\bf BUDGET 3} & \multicolumn{2}{c}{\bf BUDGET 5} \\
\multicolumn{1}{c}{\bf $\alpha$} & \multicolumn{1}{c}{\bf DEPLOYED} & \multicolumn{1}{c}{\bf MASKED} & \multicolumn{1}{c}{\bf DEPLOYED} & \multicolumn{1}{c}{\bf MASKED} \\
\hline \\
0.00 & $+0.060$ & $+0.058$ & $+0.027$ & $+0.027$ \\
0.25 & $+0.038$ & $+0.033$ & $+0.010$ & $+0.010$ \\
0.50 & $+0.014$ & $+0.048$ & $-0.000$ & $+0.040$ \\
0.75 & $-0.045$ & $+0.065$ & $-0.059$ & $+0.058$ \\
1.00 & $-0.051$ & $+0.064$ & $-0.057$ & $+0.063$ \\
\end{tabular}
\end{table}

\begin{table}[htbp]
\centering
\caption{Population-level PID of S1/S2, computed by marginalizing the $2^{22}$ background configurations. Symmetry gives equal source-specific information and hence $U_a=U_b=0$ under $I_{\min}$. Additive outcome logits at $\alpha=0$ do not imply zero PID synergy. The synergy fraction increases even though absolute synergy and joint information are non-monotonic.}
\label{tab:synergy-pid-decomposition}
\footnotesize
\setlength{\tabcolsep}{4pt}
\begin{tabular}{lrrrr}
\multicolumn{1}{c}{\bf $\alpha$} &\multicolumn{1}{c}{\bf $R$} &\multicolumn{1}{c}{\bf $\mathrm{Syn}$} &\multicolumn{1}{c}{\bf $V=R+U_a+U_b+\mathrm{Syn}$} &\multicolumn{1}{c}{\bf $\mathrm{Syn}/V$} \\
\hline \\
0.00 & $0.0471$ & $0.0505$ & $0.0976$ & $0.518$ \\
0.25 & $0.0275$ & $0.0414$ & $0.0689$ & $0.601$ \\
0.50 & $0.0117$ & $0.0637$ & $0.0754$ & $0.845$ \\
0.75 & $0.0025$ & $0.1174$ & $0.1199$ & $0.979$ \\
1.00 & $0.0000$ & $0.1951$ & $0.1951$ & $1.000$ \\
\end{tabular}
\end{table}

\begin{table}[htbp]
\centering
\caption{Acquisition rates for S1/S2, mean over five dataset instances ($n=6{,}000$ per instance and row). $P(\mathrm{both})$ is 100\% in every dataset instance at budget 8 for $\alpha\ge0.75$ but instance-dependent at $\alpha=0$ ($0\%$, $50\%$, $100\%$, $0\%$, $0\%$). At budget 3 and $\alpha=1$, S1 is abandoned in four of five instances; at budget 5 and $\alpha=1$, S2 alone consumes the entire budget in four of five.}
\label{tab:synergy-acquisition-rates}
\footnotesize
\setlength{\tabcolsep}{4pt}
\begin{tabular}{lrrrr}
\multicolumn{1}{c}{\bf $\alpha$} &\multicolumn{1}{c}{\bf BUDGET} &\multicolumn{1}{c}{\bf $P(S1)$} &\multicolumn{1}{c}{\bf $P(S2)$} &\multicolumn{1}{c}{\bf $P(\text{BOTH})$} \\
\hline \\
0.00 & 3 & 100.0\% & 0.0\% & 0.0\% \\
0.00 & 5 & 100.0\% & 0.0\% & 0.0\% \\
0.00 & 8 & 100.0\% & 29.9\% & 29.9\% \\

0.25 & 3 & 100.0\% & 0.0\% & 0.0\% \\
0.25 & 5 & 100.0\% & 0.0\% & 0.0\% \\
0.25 & 8 & 100.0\% & 50.3\% & 50.3\% \\

0.50 & 3 & 40.0\% & 0.0\% & 0.0\% \\
0.50 & 5 & 40.0\% & 0.0\% & 0.0\% \\
0.50 & 8 & 70.2\% & 80.2\% & 50.4\% \\

0.75 & 3 & 40.0\% & 0.0\% & 0.0\% \\
0.75 & 5 & 40.0\% & 60.0\% & 0.0\% \\
0.75 & 8 & 100.0\% & 100.0\% & 100.0\% \\

1.00 & 3 & 20.0\% & 0.0\% & 0.0\% \\
1.00 & 5 & 20.0\% & 80.0\% & 0.0\% \\
1.00 & 8 & 100.0\% & 100.0\% & 100.0\% \\
\end{tabular}
\end{table}

\begin{table}[htbp]
\centering
\caption{Deployed synthetic sweep: full accuracy differences against the three baselines, pooled over five dataset instances (30{,}000 paired test comparisons per cell; 10{,}000 bootstrap resamples). $^*p<0.05$ (pooled, test-level; not a between-instance test).}
\label{tab:synergy-dose-response-full}
\footnotesize
\setlength{\tabcolsep}{4pt}
\begin{tabular}{llrrr}
\multicolumn{1}{c}{\bf $\alpha$} &\multicolumn{1}{c}{\bf BASELINE} &\multicolumn{1}{c}{\bf BUDGET 3} &\multicolumn{1}{c}{\bf BUDGET 5} &\multicolumn{1}{c}{\bf BUDGET 8} \\
\hline \\
0.00 & CAE & $+0.0273^{*}$ & $+0.0102^{*}$ & $-0.0011$ \\
     & AACO & $+0.0534^{*}$ & $+0.0322^{*}$ & $+0.0194^{*}$ \\
     & Permutation & $+0.1000^{*}$ & $+0.0391^{*}$ & $+0.0041$ \\

0.25 & CAE & $+0.0182^{*}$ & $-0.0038$ & $+0.0234^{*}$ \\
     & AACO & $+0.0054$ & $-0.0054$ & $+0.0429^{*}$ \\
     & Permutation & $+0.0903^{*}$ & $+0.0392^{*}$ & $+0.0231^{*}$ \\

0.50 & CAE & $+0.0009$ & $-0.0242^{*}$ & $+0.0304^{*}$ \\
     & AACO & $-0.0104^{*}$ & $-0.0195^{*}$ & $+0.0395^{*}$ \\
     & Permutation & $+0.0509^{*}$ & $+0.0431^{*}$ & $+0.0142^{*}$ \\

0.75 & CAE & $-0.0722^{*}$ & $-0.0859^{*}$ & $+0.0687^{*}$ \\
     & AACO & $-0.0564^{*}$ & $-0.0871^{*}$ & $+0.0683^{*}$ \\
     & Permutation & $-0.0076^{*}$ & $-0.0030$ & $-0.0005$ \\

1.00 & CAE & $-0.0633^{*}$ & $-0.0903^{*}$ & $+0.1398^{*}$ \\
     & AACO & $-0.0861^{*}$ & $-0.0891^{*}$ & $+0.0917^{*}$ \\
     & Permutation & $-0.0047$ & $+0.0081^{*}$ & $+0.0000$ \\
\end{tabular}
\end{table}

\begin{table}[htbp]
\centering
\caption{Masked synthetic sweep: accuracy differences with pair proposals restricted to those affordable within the remaining budget, pooled over five dataset instances (30{,}000 paired comparisons per cell; 10{,}000 bootstrap resamples). $^*p<0.05$ (pooled, test-level).}
\label{tab:synergy-budget-mask-sensitivity-full}
\footnotesize
\setlength{\tabcolsep}{4pt}
\begin{tabular}{llrr}
\multicolumn{1}{c}{\bf $\alpha$} &\multicolumn{1}{c}{\bf BASELINE} &\multicolumn{1}{c}{\bf BUDGET 3 (MASKED)} &\multicolumn{1}{c}{\bf BUDGET 5 (MASKED)} \\
\hline \\
0.00 & CAE & $+0.0254^{*}$ & $+0.0096^{*}$ \\
     & AACO & $+0.0515^{*}$ & $+0.0316^{*}$ \\
     & Permutation & $+0.0981^{*}$ & $+0.0386^{*}$ \\

0.25 & CAE & $+0.0133^{*}$ & $-0.0039$ \\
     & AACO & $+0.0005$ & $-0.0054$ \\
     & Permutation & $+0.0854^{*}$ & $+0.0392^{*}$ \\

0.50 & CAE & $+0.0348^{*}$ & $+0.0163^{*}$ \\
     & AACO & $+0.0236^{*}$ & $+0.0210^{*}$ \\
     & Permutation & $+0.0849^{*}$ & $+0.0836^{*}$ \\

0.75 & CAE & $+0.0385^{*}$ & $+0.0303^{*}$ \\
     & AACO & $+0.0543^{*}$ & $+0.0292^{*}$ \\
     & Permutation & $+0.1032^{*}$ & $+0.1132^{*}$ \\

1.00 & CAE & $+0.0517^{*}$ & $+0.0301^{*}$ \\
     & AACO & $+0.0288^{*}$ & $+0.0313^{*}$ \\
     & Permutation & $+0.1103^{*}$ & $+0.1284^{*}$ \\
\end{tabular}
\end{table}

\begin{table}[htbp]
\centering
\caption{Fixed-$V$ calibration. Population calibration uses a $6\times10^6$-sample background draw with target $V=0.10$ bits. Realized quantities are recomputed from generated training data pooled over five dataset instances ($n=90{,}000$ each). Population symmetry implies $U_a=U_b=0$; finite-sample empirical estimates need not satisfy this exactly. Numerical small-effect checks approach $\mathrm{Syn}/V=0.5$; no general lower-bound theorem is asserted here.}
\label{tab:fixed-v-calibration}
\footnotesize
\setlength{\tabcolsep}{4pt}
\begin{tabular}{lrrrr}
\multicolumn{1}{c}{\bf TARGET $\mathrm{Syn}/V$} & \multicolumn{1}{c}{\bf $w_1{=}w_2$} & \multicolumn{1}{c}{\bf $w_{\mathrm{xor}}$} & \multicolumn{1}{c}{\bf REALIZED $V$} & \multicolumn{1}{c}{\bf REALIZED $\mathrm{Syn}/V$} \\
\hline \\
0.52 & 0.70875 & 0.0629  & 0.1032 & 0.512 \\
0.65 & 0.61239 & 0.5283  & 0.1003 & 0.644 \\
0.78 & 0.49142 & 0.73411 & 0.1003 & 0.776 \\
0.90 & 0.33509 & 0.87496 & 0.1011 & 0.896 \\
1.00 & 0.00000 & 0.97202 & 0.1017 & 1.000 \\

\end{tabular}
\end{table}

\begin{table}[htbp]
\centering
\caption{Fixed-$V$ control at budget 8: accuracy differences pooled over five independently generated dataset instances, each with its own trained policies and a classifier shared across instances (trained once, on instance 0) ($n=30{,}000$ paired test comparisons per row; 10,000 bootstrap resamples). $^*p<0.05$ describes pooled test-level uncertainty, not between-repetition significance. Instance-level trend correlations are 0.51, 0.89, 0.78, 0.85, and 0.93; all five endpoint changes are positive.}
\label{tab:fixed-v-full-significance}
\footnotesize
\setlength{\tabcolsep}{4pt}
\begin{tabular}{lrrrr}
\multicolumn{1}{c}{\bf TARGET $\mathrm{Syn}/V$} & \multicolumn{1}{c}{\bf VS.\ AACO} & \multicolumn{1}{c}{\bf VS.\ CAE} & \multicolumn{1}{c}{\bf VS.\ PERMUTATION} & \multicolumn{1}{c}{\bf MEAN} \\
\hline \\
0.52 & $+0.0207^{*}$ & $+0.0058^{*}$ & $+0.0030$ & $+0.0098$ \\
0.65 & $+0.0169^{*}$ & $+0.0310^{*}$ & $+0.0084^{*}$ & $+0.0188$ \\
0.78 & $+0.0410^{*}$ & $+0.0399^{*}$ & $+0.0081^{*}$ & $+0.0297$ \\
0.90 & $+0.0416^{*}$ & $+0.0406^{*}$ & $+0.0035^{*}$ & $+0.0286$ \\
1.00 & $+0.0431^{*}$ & $+0.0406^{*}$ & $+0.0000$ & $+0.0279$ \\
\hline
\multicolumn{5}{l}{\footnotesize Per-dataset correlation (synergy fraction vs.\ mean advantage): $0.51,\ 0.89,\ 0.78,\ 0.85,\ 0.93$} \\

\end{tabular}
\end{table}

\begin{table}[htbp]
\centering
\caption{MiniBooNE: accuracy differences ($n=26{,}014$). There are 22 significant losses, one significant win, and one non-significant comparison; $^*p<0.05$. OL-with-mask at budget 5 has $p=0.13$.}
\label{tab:miniboone-full}
\footnotesize
\setlength{\tabcolsep}{4pt}
\begin{tabular}{lrrr}
\multicolumn{1}{c}{\bf BASELINE} &\multicolumn{1}{c}{\bf BUDGET 5} &\multicolumn{1}{c}{\bf BUDGET 10} &\multicolumn{1}{c}{\bf BUDGET 15} \\
\hline \\
AACO              & $-0.034^{*}$ & $-0.053^{*}$ & $-0.033^{*}$ \\
Permutation       & $-0.038^{*}$ & $-0.047^{*}$ & $-0.049^{*}$ \\
CAE               & $-0.025^{*}$ & $-0.051^{*}$ & $-0.044^{*}$ \\
GDFS              & $-0.032^{*}$ & $-0.040^{*}$ & $-0.038^{*}$ \\
DIME              & $-0.032^{*}$ & $-0.038^{*}$ & $-0.046^{*}$ \\
JAFA              & $-0.028^{*}$ & $-0.038^{*}$ & $-0.025^{*}$ \\
OL (with mask)    & $+0.003$ ($p=0.13$) & $-0.025^{*}$ & $-0.023^{*}$ \\
OL (without mask) & $+0.018^{*}$ & $-0.036^{*}$ & $-0.026^{*}$ \\

\end{tabular}
\end{table}

\begin{table}[htbp]
\centering
\caption{Reported CKD/ACTG175 summaries ($n=80$ and 429). Low/mid/high budgets are 2/4/7 for CKD and 3/5/10 for ACTG175. ``Tie'' denotes a non-significant reported comparison; it does not imply zero effect or equivalence. This is a partial summary, not a complete eight-baseline matrix. $^*p<0.05$. For CKD ($n=80$), the three starred cells (AACO at budget 2, CAE at budget 4, permutation at budget 2) have bootstrap $p\approx0.03$ but rest on only 4, 8 and 8 discordant test instances (0 vs.\ 4 against SynAFA, 7 vs.\ 1 and 7 vs.\ 1 in its favor); an exact sign test on these pairs gives $p=0.125$, $0.070$ and $0.070$, so these three stars should be read as suggestive only. All starred ACTG175 cells remain significant under the exact test ($p\le0.009$ for DIME and JAFA, $p=0.0003$ for ODIN). EDDI denotes the external-classifier variant and ODIN the model-based variant; the built-in-classifier EDDI and model-free ODIN variants, run in the same pipeline, are significantly worse than SynAFA on ACTG175 (EDDI built-in at all three budgets, ODIN model-free at budget 10) and are omitted here.}
\label{tab:ckd-actg}
\label{tab:small-dataset-summary}
\footnotesize
\setlength{\tabcolsep}{4pt}
\begin{tabular}{llrrr}
\multicolumn{1}{c}{\bf DATASET} &\multicolumn{1}{c}{\bf BASELINE} &\multicolumn{1}{c}{\bf BUDGET (LOW)} &\multicolumn{1}{c}{\bf BUDGET (MID)} &\multicolumn{1}{c}{\bf BUDGET (HIGH)} \\
\hline \\
CKD (n=80)     & AACO        & $-0.050^{*}$ & tie & tie \\
CKD            & CAE         & tie & $+0.075^{*}$ & tie \\
CKD            & Permutation & $+0.075^{*}$ & tie & tie \\
CKD            & EDDI        & tie & tie & tie \\
CKD            & ODIN        & tie & tie & tie \\
ACTG175 (n=429)& AACO        & tie & tie & tie \\
ACTG175        & CAE         & tie & tie & tie \\
ACTG175        & Permutation & tie & tie & tie \\
ACTG175        & DIME        & $-0.028^{*}$ & $-0.037^{*}$ & tie \\
ACTG175        & JAFA        & tie & $-0.037^{*}$ & tie \\
ACTG175        & EDDI        & tie & tie & tie \\
ACTG175        & ODIN        & tie & tie & $+0.065^{*}$ \\

\end{tabular}
\end{table}

\begin{table}[htbp]
\centering
\caption{Reported BankMarketing comparisons ($n=9{,}043$). At budget 2, SynAFA, CAE and permutation select the same two features for all instances. ``Tie'' denotes the reported tie; $^*p<0.05$.}
\label{tab:bankmarketing-full}
\label{tab:bankmarketing-summary}
\footnotesize
\setlength{\tabcolsep}{4pt}
\begin{tabular}{lrrr}
\multicolumn{1}{c}{\bf BASELINE} & \multicolumn{1}{c}{\bf BUDGET 2} & \multicolumn{1}{c}{\bf BUDGET 4} & \multicolumn{1}{c}{\bf BUDGET 7} \\
\hline \\
AACO        & $+0.034^{*}$ & $-0.017^{*}$ & $-0.068^{*}$ \\
CAE         & tie & $+0.015^{*}$ & $-0.045^{*}$ \\
Permutation & tie & $+0.047^{*}$ & $-0.046^{*}$ \\

\end{tabular}
\end{table}

\begin{table}[htbp]
\centering
\caption{MiniBooNE alternative explanations. These diagnostics do not support the tested hypotheses as sufficient explanations; they do not rule out every higher-order or preprocessing effect.}
\label{tab:miniboone-hypotheses}
\footnotesize
\setlength{\tabcolsep}{4pt}
\begin{tabular}{p{0.25\linewidth}p{0.65\linewidth}}
\hline
Hypothesis & Reported diagnostic \\
\hline
Higher-order interaction & MiniBooNE's absolute three-way synergy (``Synergy3'') is second highest of five datasets (mean 0.014 bits; CKD 0.027) and mid-range relative to joint information; CKD is higher without significant losses ($n=80$, low power). This does not test all higher orders. \\
Binarization loss & The reported retained-information fraction is 59.5\% (second lowest of five datasets), versus 40.7\% for PhysioNet, which performs better. \\
Training sample size & The AACO--SynAFA gap (budget 10) does not shrink as the training set grows in a subsampling sweep from $n=300$ (0.067) to 78,038 (0.044). \\
\hline
\end{tabular}
\end{table}

\begin{table}[htbp]
\centering
\caption{MiniBooNE classifier realignment: acquisition trajectories are fixed while the external classifier is retrained using 96,000 policy masks from training-split rollouts. Entries are SynAFA minus baseline accuracy ($n=26{,}014$). Closed is the percentage reduction in absolute gap, averaged over the nine rows for the final mean. $^*p<0.05$.}
\label{tab:miniboone-classifier-realignment}
\footnotesize
\setlength{\tabcolsep}{4pt}
\begin{tabular}{lcccc}
\multicolumn{1}{c}{\bf BASELINE} &
\multicolumn{1}{c}{\bf BUDGET} &
\multicolumn{1}{c}{\bf ORIGINAL} &
\multicolumn{1}{c}{\bf POLICY-ALIGNED} &
\multicolumn{1}{c}{\bf CLOSED}
\\ \hline \\
AACO        & 5  & $-0.0343^{*}$ & $-0.0007$ (n.s.) & 98\% \\
CAE         & 5  & $-0.0252^{*}$ & $-0.0176^{*}$    & 30\% \\
Permutation & 5  & $-0.0381^{*}$ & $-0.0200^{*}$    & 47\% \\
AACO        & 10 & $-0.0525^{*}$ & $-0.0222^{*}$    & 58\% \\
CAE         & 10 & $-0.0506^{*}$ & $-0.0296^{*}$    & 41\% \\
Permutation & 10 & $-0.0469^{*}$ & $-0.0305^{*}$    & 35\% \\
AACO        & 15 & $-0.0335^{*}$ & $-0.0121^{*}$    & 64\% \\
CAE         & 15 & $-0.0444^{*}$ & $-0.0361^{*}$    & 19\% \\
Permutation & 15 & $-0.0492^{*}$ & $-0.0350^{*}$    & 29\% \\
\hline
{\bf Mean} & & & & {\bf 47\%} \\
\end{tabular}
\end{table}

\begin{table}[htbp]
\centering
\caption{Reduced $D=90$ control, adaptive acquisition without local search ($n=4{,}000$). Gaps reproduce the reported exact-minus-amortized differences and may differ from subtraction of rounded accuracies. Fit times do not establish total end-to-end precomputation cost. $p$-values come from a paired bootstrap over test instances with 5{,}000 resamples.}
\label{tab:amortized-k-infer}
\footnotesize
\setlength{\tabcolsep}{4pt}
\begin{tabular}{lrrl}
\multicolumn{1}{c}{\bf CONFIGURATION} &\multicolumn{1}{c}{\bf ACCURACY} &\multicolumn{1}{c}{\bf GAP TO EXACT} &\multicolumn{1}{c}{\bf FIT TIME} \\
\hline \\
Exact pairwise information       & 0.7632 & --      & -- \\
Amortized, k\_infer=150 (original)  & 0.7505 & $+0.0128^{*}$ ($p=0.02$) & 25.6s \\
Amortized, k\_infer=1000           & 0.7538 & $+0.0095$ ($p=0.08$)     & 24.3s \\
Amortized, k\_infer=36{,}000 (full) & 0.7628 & $+0.0005$ ($p=0.91$)    & 24.8s \\

\end{tabular}
\end{table}

\begin{table}[htbp]
\centering
\caption{MNIST-loop accuracy on the full 12,000-instance test set across amortized configurations. Increasing the precompute sample and correcting replacement sampling are separate changes.}
\label{tab:d784-full}
\footnotesize
\setlength{\tabcolsep}{4pt}
\begin{tabular}{lrrr}
\multicolumn{1}{c}{\bf CONFIGURATION} &\multicolumn{1}{c}{\bf BUDGET 5} &\multicolumn{1}{c}{\bf BUDGET 10} &\multicolumn{1}{c}{\bf BUDGET 20} \\
\hline \\
Original (k\_infer=150)                    & 0.695 & 0.786 & \textbf{0.814} \\
k\_infer=36{,}000, with-replacement sampling & 0.719 & 0.770 & 0.811 \\
k\_infer=36{,}000, seed=1 (variance check)  & 0.752 & 0.754 & 0.800 \\
k\_infer=36{,}000, extended training $k$-range (rejected) & 0.735 & 0.765 & 0.790 \\
k\_infer=36{,}000, \textbf{sampling corrected} & \textbf{0.748} & 0.776 & 0.805 \\

\end{tabular}
\end{table}

\begin{table}[htbp]
\centering
\caption{MNIST-loop local-search diagnostic at $D=784$, budget 20, on the sampling-corrected amortized table (\texttt{k\_infer}=36{,}000). The first row uses a random subsample of $n=1000$ test instances (the setting of the original diagnostic); the second uses the full 12{,}000-instance test set, whose adaptive accuracy matches Table~\ref{tab:d784-full}. Mean $T(S)$ before/after search: $2.60\to3.77$ ($n=1000$) and $2.56\to3.74$ (full).}
\label{tab:local-search-progression}
\footnotesize
\setlength{\tabcolsep}{3pt}
\begin{tabular}{lrrr}
& \multicolumn{1}{c}{\bf ADAPTIVE ACC.} &\multicolumn{1}{c}{\bf REVISED ACC.} &\multicolumn{1}{c}{\bf \% CAE GAP CLOSED} \\
\hline
Corrected search, $n=1000$ subsample & 0.836 & 0.806 & $-$73.2\% (reversed) \\
Corrected search, full test set ($n=12{,}000$) & 0.805 & 0.773 & $-$46.3\% (reversed) \\
\end{tabular}
\end{table}

\clearpage
\begin{landscape}
\begin{table}[p]
\centering
\caption{Single-swap revision (corrected local search under the diagnostic $T(S)$ objective): revised minus adaptive accuracy, evaluated on the \emph{full} test set of each dataset (MNIST-loop $n=12{,}000$, PhysioNet $n=2{,}400$, MiniBooNE $n=26{,}014$) with the shared external classifier; $\lambda$ is each trained bundle's own value ($0.001$ for MNIST-loop, $0.01$ otherwise) and every instance converged within 200 rounds. 95\% CI and $p$ come from a paired bootstrap over test instances (10{,}000 resamples; $p<0.0001$ means no resample crossed zero). None of the nine settings shows a gain, and seven show significant losses; $^*$: $p<0.05$. Below the line, the reduced-dimensional exact control (exact PID on the 90 highest-variance pixels, $\lambda=0.001$, same test set and bootstrap) improves accuracy at budgets 5 and 20 but reduces it at budget 10.}
\label{tab:local-search-generalization}
\footnotesize
\setlength{\tabcolsep}{3pt}
\begin{tabular}{llrrcl}
\multicolumn{1}{c}{\bf SETTING} & \multicolumn{1}{c}{\bf BUDGET} & \multicolumn{1}{c}{\bf $n$} & \multicolumn{1}{c}{\bf $\Delta$ ACC.} & \multicolumn{1}{c}{\bf 95\% CI} & \multicolumn{1}{c}{\bf $p$} \\
\hline
MNIST-loop (amortized, sampling-corrected) & 5 & 12{,}000 & $-0.0025$ & [$-0.0106$, $+0.0054$] & 0.547 \\
MNIST-loop (amortized, sampling-corrected) & 10 & 12{,}000 & $-0.0213^{*}$ & [$-0.0288$, $-0.0138$] & $<0.0001^{*}$ \\
MNIST-loop (amortized, sampling-corrected) & 20 & 12{,}000 & $-0.0322^{*}$ & [$-0.0400$, $-0.0244$] & $<0.0001^{*}$ \\
PhysioNet (exact) & 5 & 2{,}400 & $-0.0371^{*}$ & [$-0.0512$, $-0.0221$] & $<0.0001^{*}$ \\
PhysioNet (exact) & 10 & 2{,}400 & $-0.0150^{*}$ & [$-0.0275$, $-0.0025$] & $0.019^{*}$ \\
PhysioNet (exact) & 15 & 2{,}400 & $-0.0042$ & [$-0.0158$, $+0.0079$] & 0.514 \\
MiniBooNE (exact) & 5 & 26{,}014 & $-0.0404^{*}$ & [$-0.0442$, $-0.0366$] & $<0.0001^{*}$ \\
MiniBooNE (exact) & 10 & 26{,}014 & $-0.0148^{*}$ & [$-0.0181$, $-0.0113$] & $<0.0001^{*}$ \\
MiniBooNE (exact) & 15 & 26{,}014 & $-0.0065^{*}$ & [$-0.0099$, $-0.0033$] & $<0.0001^{*}$ \\
\hline
\emph{For reference: exact PID on a reduced $D{=}90$ pixel pool} & & & & & \\
MNIST-loop, $D{=}90$ pixel pool (exact) & 5 & 12{,}000 & $+0.0767^{*}$ & [$+0.0670$, $+0.0866$] & $<0.0001^{*}$ \\
MNIST-loop, $D{=}90$ pixel pool (exact) & 10 & 12{,}000 & $-0.0125^{*}$ & [$-0.0200$, $-0.0052$] & $0.0002^{*}$ \\
MNIST-loop, $D{=}90$ pixel pool (exact) & 20 & 12{,}000 & $+0.0164^{*}$ & [$+0.0108$, $+0.0220$] & $<0.0001^{*}$ \\
\end{tabular}
\end{table}
\end{landscape}
\clearpage

\subsection{Synthetic sweep: generative details}
\label{app:synthetic-generative}
Every synthetic dataset instance has 30{,}000 samples, split 60/20/20 into train/validation/test (18{,}000/6{,}000/6{,}000), the default split ratio used for all datasets in this paper. The 24 features are independent Bernoulli$(0.5)$: 22 background features $B_1,\dots,B_{22}$ (cost 1 each) and the designated pair S1 (cost 1) and S2 (cost 5). The outcome is $Y\sim\mathrm{Bernoulli}(\sigma(\ell))$ with logit
\[
\ell=\sum_{i=1}^{22} b_i(2B_i-1) + (1-\alpha)\big[w_1(2S_1-1)+w_2(2S_2-1)\big] + \alpha\, w_{\mathrm{xor}}\big[2(S_1 \oplus S_2)-1\big],
\]
with no intercept (every additive term uses the symmetric $\pm1$ encoding, so the logit has zero mean and the classes are approximately balanced), $w_1=w_2=0.7$, $w_{\mathrm{xor}}=1.4$, and background weights $b=(0.50,\allowbreak 0.46,\allowbreak 0.42,\allowbreak 0.38,\allowbreak 0.34,\allowbreak 0.30,\allowbreak 0.27,\allowbreak 0.24,\allowbreak 0.21,\allowbreak 0.18,\allowbreak 0.16,\allowbreak 0.14,\allowbreak 0.12,\allowbreak 0.10,\allowbreak 0.09,\allowbreak 0.08,\allowbreak 0.07,\allowbreak 0.06,\allowbreak 0.05,\allowbreak 0.04,\allowbreak 0.03,\allowbreak 0.02)$. In the fixed-$V$ control the mixture is replaced by $w_{\mathrm{add}}[(2S_1-1)+(2S_2-1)]+w_{\mathrm{xor}}[2(S_1\oplus S_2)-1]$ with $(w_{\mathrm{add}},w_{\mathrm{xor}})$ from Table~\ref{tab:fixed-v-calibration}.

Dataset instances $0,\dots,4$ are generated (data and split) with seeds $42,\dots,46$. Both the primary $\alpha$-sweep (Tables~\ref{tab:synergy-dose-response} and \ref{tab:synergy-dose-response-full}) and the fixed-$V$ control use all five instances $0,\dots,4$; the main $\alpha$-sweep table reports the mean over instances (standard deviation across instances in parentheses) and the full tables pool the 30{,}000 paired test comparisons per cell. With only five instances, the between-instance correlations are descriptive. For every synthetic dataset the external classifier is trained once, on the training and validation data of instance 0, and reused for all instances; SynAFA, CAE, AACO and the permutation baseline are trained separately for each instance (training seed equal to the instance index).

\subsection{External classifier: architecture and training}
\label{app:external-classifier}
The external classifier used for the primary comparisons is a masked MLP: two hidden layers of 128 units each (ReLU, dropout 0.1-0.2), taking as input the concatenation of the masked feature vector (zeros for unobserved features) and a binary feature mask (1 if observed, 0 if not), so the input dimension is $2D$. It is trained once per dataset with class-weighted cross-entropy (Adam, lr$=10^{-3}$; 100 epochs for CKD, ACTG175, PhysioNet and the synthetic datasets, 300 for BankMarketing, 1{,}000 for MiniBooNE and 60 for MNIST-loop; the checkpoint with the lowest validation loss under the minimum masking probability is kept, i.e., with all features observed for the tabular datasets) and reused, unchanged, for every method and every budget compared on that dataset.

Training uses a random feature-masking curriculum: in every batch, a fraction of features is zeroed out, with the masking probability drawn uniformly from a dataset-specific range (0.0--0.9 for the five tabular datasets; 0.75--0.99 for MNIST-loop, reflecting the much smaller fraction of $D=784$ pixels any tested budget actually acquires). This is what lets a single fixed classifier give a meaningful prediction from whatever partial feature subset a given acquisition policy selects, at any budget, without being retrained per method or per budget.

\subsection{MiniBooNE: alternative explanations, acquisition collapse, and classifier sensitivity}
\label{app:miniboone}

Table~\ref{tab:miniboone-hypotheses} summarizes three mechanistic hypotheses for MiniBooNE's negative pattern (22 significant losses, one significant win, and one non-significant comparison across eight baselines and three budgets), each tested with a direct, exact measurement rather than assumed. \textbf{Higher-order interaction}: we computed exact 3-way synergy (``Synergy3'') over all feature triples on every dataset; MiniBooNE's mean, 90th-percentile and maximum values (0.014, 0.032 and 0.065 bits) are the second highest of the five datasets in absolute terms, behind only CKD (0.027, 0.061 and 0.122 bits) and far above the other three (mean at most 0.003 bits), while relative to joint information they are mid-range (rank 3 of 5); CKD has more 3-way synergy but shows no significant losses (its 80-instance test set has little power). These diagnostics therefore do not clearly support higher-order interaction as the differentiator, but they cannot exclude it. \textbf{Binarization loss}: we measured the fraction of continuous mutual information recovered after median-split binarization on every dataset; MiniBooNE recovers 59.5\%, the second lowest of the five datasets (CKD 67.5\%, ACTG175 70.6\%, BankMarketing 60.2\%, PhysioNet 40.7\%); PhysioNet, the only dataset with a lower fraction, is one where SynAFA performs comparatively well, the opposite of what this hypothesis predicts for MiniBooNE's failures. Binarization loss is therefore not shown to be the differentiator, although a contribution cannot be excluded. \textbf{Sample size}: a controlled subsampling sweep of MiniBooNE's training set, from $n=300$ to the full 78,038, shows that the AACO$-$SynAFA accuracy gap at budget 10 does not shrink as the training set grows: it is 0.067, 0.037, 0.024, 0.044 and 0.044 at $n=300$, 1{,}500, 8{,}000, 30{,}000 and 78{,}038 (one training run per size, no monotonic trend), including at $n=300$, where MiniBooNE is no longer the largest dataset in the comparison. These diagnostics do not support sample size as the explanation.

\textbf{Observed acquisition collapse: state-independent selection.} Directly inspecting SynAFA's acquisition trajectories on MiniBooNE's full test set ($n=26{,}014$) reveals that SynAFA acquires the \emph{identical} feature subset for every single instance, at every tested budget (5, 10, and 15) -- exactly one unique acquisition pattern across all 26,014 instances, at each budget. This is unique to MiniBooNE among the five tabular datasets: the same check finds 2 unique patterns on CKD ($n=80$, budget 4), 5 on ACTG175 ($n=429$, budget 5), 3 on BankMarketing ($n=9{,}043$, budget 4), and 10 on PhysioNet ($n=2{,}400$, budget 10) -- with the majority pattern covering 50--82\% of instances rather than 100\%. This is not a data artifact: MiniBooNE's features vary substantially instance to instance (998 of 1,000 sampled training feature vectors are pairwise distinct).

We traced this to the acquisition rule directly. \S\ref{sec:method}'s single-feature score is $\mathrm{contrib}(j\mid S) = \max(\texttt{single\_value}[j], \max_{k \in S}\texttt{cond\_single\_value}[j,k,x_k])$; only the second, conditional term depends on instance-specific realized values $x_k$. At the first branching decision on MiniBooNE (choosing the 6th of 10 features, with 5 already acquired; every test instance shares this state), we exhaustively enumerated all $2^5=32$ possible realized-value combinations for the acquired features: the winning proposal was the same in every combination (feature 1). It is an unstarted-pair proposal, whose score (0.171) does not depend on realized values and exceeds the best single-feature proposal (0.104) by 0.066; even within the single-feature branch alone, the top-ranked candidate was identical across all 32 combinations, with a margin (best minus second-best single-feature score) never below 0.008. The same exhaustive check on PhysioNet, at the 6th decision of budget 10 for each of the six acquisition prefixes that occur there (together covering all test instances), found two or three different winning features depending on realized values, with minimum margins to a proposal for a different feature between 0.0000 and 0.0007 (0.0001--0.0011 within the single-feature branch). The difference is one of scale: MiniBooNE's population-level scores (maximum \texttt{single\_value} 0.213 vs.\ 0.028 for PhysioNet, and the state-independent pair scores above) dominate the realized-value-dependent conditional terms at the inspected decision, whereas PhysioNet's competing proposals are close enough for realized values to change the choice. This supports score dominance at that decision without proving invariance at every possible unvisited state; the observation that all 26{,}014 test trajectories select the same subset covers the states actually visited. SynAFA's state-dependent branch therefore never actually changes which features are acquired on MiniBooNE: the policy collapses to a fixed, population-level, one-size-fits-all selection, identical for every instance.

This identifies a mechanism for the observed lack of adaptation, without establishing its causal contribution to the accuracy gap. Among the AACO/CAE/permutation comparison subset used for this trajectory diagnostic, AACO retains substantial personalization on MiniBooNE (5,398 unique acquisition patterns across the same 26,014 test instances at budget 10, vs.\ SynAFA's one); CAE and permutation are static-selection baselines whose acquisitions we verified are instance-invariant by construction on every dataset we checked, including PhysioNet, so their invariance here is not informative about SynAFA's mechanism specifically. We stop short of claiming this fully accounts for the observed accuracy gap's magnitude. We have not established a direct counterfactual (e.g., a variant forced to diversify its acquisitions) proving that restoring adaptivity would close it, but a policy that cannot personalize forgoes exactly the kind of instance-specific signal AACO is free to exploit.

\textbf{A second, partially confounding factor: evaluation-harness distribution shift.} Even granting a fixed, instance-invariant acquisition, is the reported accuracy gap a faithful measure of that acquisition's quality, or is it inflated by which masks the shared external classifier happened to see during its own training? We tested this directly. The external classifier is trained by masking each feature independently at random (Appendix~\ref{app:external-classifier}); we collected real acquisition masks from SynAFA and the three main comparison baselines (AACO, CAE, permutation) on MiniBooNE's training split (8,000 instances, all three budgets), pooled them into an empirical mask bank (96,000 rows), and retrained a second external classifier under an otherwise-identical curriculum that samples masks from this bank instead of i.i.d.\ Bernoulli masking. Re-evaluating all four methods on the real test set ($n=26{,}014$) with this policy-aligned classifier, holding every acquisition trajectory fixed, closes SynAFA's accuracy gap to AACO/CAE/permutation by a mean of 47\% across the nine budget$\times$baseline comparisons, leaving a non-significant residual gap to AACO at budget 5, with baseline-dependent changes in their own accuracies (Table~\ref{tab:miniboone-classifier-realignment}). SynAFA's accuracy improves at every budget (+1.7, +2.3, and +0.9 percentage points at budgets 5, 10, and 15). The other policies change by at most 0.9 points except AACO, the adaptive comparator in this four-policy classifier control, whose accuracy decreases by 0.7--1.7 points, so part of the closed gap to AACO reflects AACO's decrease rather than SynAFA's gain. This demonstrates sensitivity of the relative gap to classifier mask training. It does not establish that a particular mask was never seen, or that every non-adaptive policy suffers the same shift. SynAFA still has significant losses in 8 of 9 comparisons after realignment, but the cause of the residual gap is not isolated by this control. \S\ref{sec:pattern}'s budget-erosion pattern remains a different, independently-supported finding and is not offered as an explanation for MiniBooNE specifically.

\subsection{Local search: full methodology}
\label{app:local-search}
Let $S$ be the current set of acquired features for a given test
instance, with realized values $x$. Define each acquired feature
$j \in S$'s contribution as
\[
\mathrm{contrib}(j \mid S)
=
\max\left\{
\texttt{single\_value}[j],
\max_{k \in S\setminus\{j\}}
\texttt{cond\_single\_value}[j,k,x_k]
\right\}.
\]
The total set score is
\[
T(S)
=
\sum_{j \in S}
\mathrm{contrib}(j \mid S).
\]

\textbf{Uncorrected version.} Each round: let $w = \arg\min_{j \in S} \mathrm{contrib}(j \mid S)$ (the current worst member) and $c^* = \arg\max_{c \notin S} \mathrm{contrib}(c \mid S)$ (the best outside candidate, scored against the \emph{unmodified} $S$, i.e., still including $w$). Swap $w \to c^*$ if $\mathrm{contrib}(c^* \mid S) > \mathrm{contrib}(w \mid S) + \epsilon$. This does not track $T(S)$ correctly: removing $w$ can change other members' best-partner conditioning (if $w$ was their best partner) and $c^*$'s own value was scored while $w$ was still present, so accepted ``improving'' swaps do
not provably increase the true $T(S)$.

\textbf{Corrected version.} Each round, for every candidate removal position $j \in S$: let $B = S \setminus \{j\}$, and find $c^* =
\arg\max_{c \notin B} T(B \cup \{c\})$ by exactly recomputing $T(B \cup \{c\})$ for every candidate $c$ (vectorized: $\mathrm{contrib}(m \mid
B\cup\{c\})$ for existing members $m \in B$ can change because $c$ is a new potential best partner, and this is computed exactly for all $c$
simultaneously via one matrix operation over \texttt{cond\_single\_value}). Accept the single best $(j, c^*)$ pair found across all removal positions if it strictly increases $T(S)$. Because the number of candidate feature subsets is finite and $T(S)$ strictly increases after every accepted swap, the procedure must terminate after finitely many accepted moves. Empirically, all instances converged within 200 rounds.

\paragraph{Diagnostic search procedure.}
For each instance, initialize $S$ to the deployed policy's acquired set and repeat:
\begin{enumerate}
\item For every $j\in S$ and $c\notin S$, form $S'=(S\setminus\{j\})\cup\{c\}$ and recompute every summand of $T(S')$, including the effect of $c$ on existing members.
\item Choose a candidate with the largest $T(S')$. If its value is not strictly greater than $T(S)$, terminate; otherwise replace $S$ with that candidate and repeat.
\end{enumerate}
This diagnostic keeps set cardinality fixed in the uniform-cost settings studied here. It evaluates realized values of candidate replacements retrospectively, without charging the cumulative cost of discarded measurements, and therefore is not itself an online hard-budget policy. Evaluating the unchanged set as an additional candidate is equivalent when acceptance requires a strict increase.

\textbf{Generalization test (\S\ref{sec:local-search-test}).} We then evaluated the corrected local-search procedure across nine settings: all 3 MNIST-loop budgets using the $D=784$ amortized table, and all 3 budgets on PhysioNet ($D=41$) and MiniBooNE ($D=50$) using the exact tables. In each case, the starting set $S_0$ was the real adaptive policy's own selection (the one actually deployed), so the comparison tests the value of allowing revision under this diagnostic objective. Table~\ref{tab:local-search-generalization} reports all nine results (plus the exact-scoring $D=90$ control for reference). Local search never significantly improved accuracy in any of the nine settings, and was significantly worse in seven (MNIST-loop budgets 10 and 20, PhysioNet budgets 5 and 10, MiniBooNE budgets 5, 10 and 15). All nine settings use the full test sets, with a paired bootstrap over test instances (10{,}000 resamples) for the accuracy change. Mean swaps per instance ranged from 1.8 to 10.9 (4{,}225 to 131{,}317 accepted swaps per setting), and every instance converged within 200 rounds, confirming that the lack of performance improvement was not due to an absence of accepted revisions.

\textbf{Objective-alignment analysis (\S\ref{sec:objective-alignment}).}
For each of the 553{,}702 accepted swaps across the nine settings, we recorded the pre- and post-swap feature masks and batch-scored both through the same external classifier, reading off the predicted probability of the true class as $U$. $\Delta U$ is this probability's change; a random-swap baseline uses the same removal position with a uniformly random replacement candidate instead of local search's chosen one, giving a same-instance, same-removal control for whether $T(S)$'s specific choice does better than a random replacement. Full per-setting results, including the random-swap baseline hit rates, are in Table~\ref{tab:objective-alignment}. Correlations are Pearson's; confidence intervals use an instance-level cluster bootstrap (10{,}000 resamples), and Spearman correlations range from $-0.077$ to $+0.168$. All per-instance and per-swap outputs are stored with the code.

\clearpage
\begin{landscape}
\begin{table}[p]
\centering
\caption{Alignment between the diagnostic set-level objective $T(S)$ and predictive utility, from the same full-test-set runs as Table~\ref{tab:local-search-generalization}. Every accepted swap increases $T(S)$ by construction. $\Delta U$ is the change in the external classifier's true-class probability for the swap chosen by $T(S)$; the random control keeps the same removal and draws a uniformly random replacement. \#Swaps is the number of accepted swaps. $\rho$ is Pearson's correlation of $\Delta T$ and $\Delta U$ over accepted swaps, with 95\% CIs from an instance-level cluster bootstrap (10{,}000 resamples; swaps from one instance are dependent). $|\rho|\le0.11$ in the nine main settings; the last rows show the exact-scoring $D{=}90$ control, whose correlation is largest ($+0.17$) at budget 5, where revision helps.}
\label{tab:objective-alignment}
\footnotesize
\setlength{\tabcolsep}{3pt}
\begin{tabular}{llrrrrrl}
\multicolumn{1}{c}{\bf SETTING} & \multicolumn{1}{c}{\bf BUDGET} & \multicolumn{1}{c}{\bf $n$} & \multicolumn{1}{c}{\bf \#SWAPS} & \multicolumn{1}{c}{\bf MEAN $\Delta U$} & \multicolumn{1}{c}{\bf $P(\Delta U>0)$} & \multicolumn{1}{c}{\bf $P(\Delta U_{\rm rand}>0)$} & \multicolumn{1}{c}{\bf $\rho(\Delta T,\Delta U)$ [95\% CI]} \\
\hline
MNIST-loop & 5 & 12{,}000 & 45{,}189 & $+0.0014$ & 0.506 & 0.400 & $+0.073$ [$+0.063$, $+0.082$] \\
MNIST-loop & 10 & 12{,}000 & 79{,}544 & $+0.0007$ & 0.520 & 0.426 & $+0.107$ [$+0.099$, $+0.114$] \\
MNIST-loop & 20 & 12{,}000 & 131{,}317 & $-0.0024$ & 0.485 & 0.418 & $+0.044$ [$+0.038$, $+0.050$] \\
PhysioNet & 5 & 2{,}400 & 4{,}225 & $-0.0116$ & 0.450 & 0.422 & $-0.010$ [$-0.040$, $+0.019$] \\
PhysioNet & 10 & 2{,}400 & 5{,}386 & $-0.0022$ & 0.482 & 0.464 & $-0.068$ [$-0.094$, $-0.042$] \\
PhysioNet & 15 & 2{,}400 & 8{,}903 & $+0.0004$ & 0.533 & 0.535 & $-0.048$ [$-0.072$, $-0.024$] \\
MiniBooNE & 5 & 26{,}014 & 61{,}571 & $-0.0180$ & 0.419 & 0.400 & $+0.095$ [$+0.088$, $+0.103$] \\
MiniBooNE & 10 & 26{,}014 & 99{,}614 & $-0.0038$ & 0.498 & 0.504 & $+0.060$ [$+0.056$, $+0.065$] \\
MiniBooNE & 15 & 26{,}014 & 117{,}953 & $-0.0039$ & 0.453 & 0.468 & $+0.006$ [$+0.002$, $+0.011$] \\
\hline
\emph{For reference: exact PID, $D{=}90$ pool} & & & & & & & \\
MNIST-loop, $D{=}90$ pixel pool (exact) & 5 & 12{,}000 & 17{,}316 & $+0.0151$ & 0.490 & 0.446 & $+0.170$ [$+0.154$, $+0.185$] \\
MNIST-loop, $D{=}90$ pixel pool (exact) & 10 & 12{,}000 & 34{,}315 & $+0.0007$ & 0.503 & 0.479 & $-0.033$ [$-0.044$, $-0.023$] \\
MNIST-loop, $D{=}90$ pixel pool (exact) & 20 & 12{,}000 & 41{,}163 & $+0.0025$ & 0.509 & 0.494 & $-0.035$ [$-0.045$, $-0.026$] \\
\end{tabular}
\end{table}
\end{landscape}
\clearpage

\subsection{Amortized estimator: implementation and ablations}
\label{app:amortized}

Exact computation of the pairwise information tables requires $O(D^2)$ feature-pair evaluations, each estimated from $O(n)$ training samples, which is tractable for moderate $D$ but expensive at image scale ($D=784$, yielding $\binom{784}{2}=306{,}936$ pairs). Marginal information $M_j$ requires only $O(D)$ evaluations and is therefore computed exactly at every $D$, including $D=784$. Only the two pairwise tables, $V_{jk}$ and $C_{j\mid k,v}$, are estimated. To scale SynAFA to this setting, we introduce an amortized cross-pair estimator that learns to predict pairwise information values from small empirical samples. Specifically, we use a hurdle model consisting of a classifier that predicts whether a pair exceeds a fixed non-triviality threshold and a regressor that estimates its magnitude conditional on exceeding that threshold. Each model uses its own feature vector, built from its own $k$-row sample. For $V_{jk}$, an eight-cell empirical joint frequency of $(X_j,X_k,Y)$ plus $k$, from a $k$-row sample drawn unconditionally. For $C_{j\mid k,v}$, a four-cell empirical joint frequency of $(X_j,Y)$ plus $k$, from a $k$-row sample drawn only from rows with $X_k=v$. By training across many reference pairs, the estimator shares statistical information across pairwise frequency patterns and can estimate each value from $k \ll n$ samples rather than recomputing it from the full training set.

Both \texttt{pair\_value}$[j,k]$ and \texttt{cond\_single\_value}$[j,k,v]$ depend only on the training data and feature indices, not on the test instance. We therefore precompute both tables offline: $V_{jk}$ from one fixed sample of the whole training set, and $C_{j\mid k,v}$ from a fixed sample per $(k,v)$ stratum, both via vectorized matrix operations. At test time, SynAFA v2 reads only information table entries, so the acquisition-time computation is identical to the exact version.

\textbf{Hurdle model.}
A \texttt{RandomForestClassifier} (300 trees) predicts whether a pair's joint-information value exceeds a fixed non-triviality threshold (0.002 bits), and a \texttt{RandomForestRegressor} (300 trees), fitted only on the non-trivial subset, predicts its magnitude. The final estimate is
\[
\hat{V}
=
\mathbf{1}[\hat{p} \ge \tau]\hat{V}_{\mathrm{reg}},
\]
where $\tau$ is selected on a held-out split of reference pairs to
minimize MAE against exact joint-information values.

\textbf{Estimating the conditional table.}
$\texttt{cond\_single\_value}[j,k,v] = I(Y;X_j \mid X_k{=}v)$ is estimated by a separate hurdle model (its own classifier and regressor, same architecture and 0.002-bit non-triviality cutoff, but its own threshold, reselected on its own held-out split) trained on reference examples that target the exact conditional value. For each reference pair, $v=0$ and $v=1$ examples are pooled into one shared training set rather than fit as separate per-value models; each example is built from a random $k$-row draw ($k$ log-uniform in $[20,300]$, with 4{,}000 reference pairs and 8 draws per pair) taken only from rows where $X_k{=}v$.

\textbf{Precomputation sampling correction.}
During estimator diagnostics, we found that the original precomputation procedure sampled with replacement even when
\texttt{k\_infer} was at least as large as the training-set size. When drawing $n$ samples from $n$ training instances with replacement, approximately 36.8\% of instances are omitted on average while others are sampled repeatedly. We corrected the procedure so that, when the requested sample size covers the full training population, every training instance is used exactly once. At \texttt{k\_infer}$=n$, this also makes precomputation deterministic.
Table~\ref{tab:d784-full} separates increasing sample size from correcting sampling. At budget 5, accuracy is 0.695 for the original \texttt{k\_infer}$=150$, 0.719 for \texttt{k\_infer}$=36{,}000$ with replacement, and 0.748 for the corrected full-sample configuration. The previously reported CAE-gap change from $-0.077$ to $-0.023$ combines the first and last configurations, and cannot be attributed solely to the sampling correction. Larger-budget gaps remain.

\textbf{Training-range ablation.}
Because inference with \texttt{k\_infer}$=36{,}000$ uses substantially larger empirical samples than the $k\in[20,300]$ reference samples used to train the hurdle model, we tested whether extending the training range to larger $k$ would improve estimation at image scale. It did not: performance decreased rather than improved, with the largest reduction at budget 20 ($0.811$ to $0.790$; Table~\ref{tab:d784-full}). Thus, simply expanding the training $k$-range did not resolve the high-dimensional approximation gap.

\subsection{Pair-branch ablation}
\label{app:pair-off}
To test whether the effects attributed to pairwise information depend on SynAFA's unstarted-pair proposal branch, we re-ran the decision loop with that branch removed: only the single-feature score $\max\{M_j,\max_k C_{j\mid k,x_k}\}-\lambda c_j$ remains, with the same information tables, external classifier, budgets, $\lambda$ and forced-acquisition fallback. The loop is the re-implementation used for the feasibility mask; with the pair branch present it reproduces the official results (exactly in all 75 synthetic instance-by-$\alpha$-by-budget cells, and with identical predictions on the first 400 test instances of each of the 15 real-data settings). This control removes pair proposals but keeps conditioning on realized values.\par\smallskip\noindent

\begin{table}[htbp]
\centering
\caption{Pair-branch ablation on the synthetic $\alpha$-sweep: mean accuracy advantage over CAE, AACO and permutation (five dataset instances) for the deployed policy and with the pair branch removed (Off). The last column is the paired accuracy difference Deployed$-$Off at budget 8, pooled over the 30{,}000 test comparisons ($^*p<0.05$, test-level, not a between-instance test). The last row is the correlation with $\alpha$.}
\label{tab:pair-off-alpha}
\footnotesize
\setlength{\tabcolsep}{3pt}
\begin{tabular}{lrrrrrrr}
$\alpha$ & \multicolumn{2}{c}{Budget 3} & \multicolumn{2}{c}{Budget 5} & \multicolumn{2}{c}{Budget 8} & Branch effect (B8) \\
 & Deployed & Off & Deployed & Off & Deployed & Off & (Dep.$-$Off acc.) \\
\hline
0.00 & +0.060 & +0.058 & +0.027 & +0.026 & +0.007 & +0.006 & $+0.001$ \\
0.25 & +0.038 & +0.033 & +0.010 & +0.005 & +0.030 & +0.030 & $-0.000$ \\
0.50 & +0.014 & +0.048 & -0.000 & +0.040 & +0.028 & +0.001 & $+0.027^{*}$ \\
0.75 & -0.045 & +0.065 & -0.059 & +0.058 & +0.045 & -0.005 & $+0.051^{*}$ \\
1.00 & -0.051 & +0.064 & -0.057 & +0.063 & +0.077 & -0.043 & $+0.120^{*}$ \\
\hline
corr. with $\alpha$ & -0.97 & +0.51 & -0.95 & +0.85 & +0.95 & -0.80 & \\
\end{tabular}
\end{table}

\begin{table}[htbp]
\centering
\caption{Pair-branch ablation on the fixed-$V$ sweep (budget 8, five dataset instances pooled): mean advantage over CAE, AACO and permutation, deployed versus pair branch removed, and the pooled paired accuracy difference ($^*p<0.05$, test-level).}
\label{tab:pair-off-fixedv}
\footnotesize
\setlength{\tabcolsep}{3pt}
\begin{tabular}{rrrr}
Realized Syn/$V$ & Deployed & Off & Branch effect (Dep.$-$Off acc.) \\
\hline
0.512 & +0.0098 & +0.0100 & $-0.0002$ \\
0.644 & +0.0188 & +0.0328 & $-0.0140^{*}$ \\
0.776 & +0.0297 & +0.0419 & $-0.0122^{*}$ \\
0.896 & +0.0286 & -0.0087 & $+0.0373^{*}$ \\
1.000 & +0.0279 & +0.0032 & $+0.0246^{*}$ \\
\hline
corr. with Syn/$V$ & +0.874 & -0.395 & \\
\end{tabular}
\end{table}

\begin{table}[htbp]
\centering
\caption{Pair-branch ablation on the real datasets: test accuracy of deployed SynAFA and with the pair branch removed, their paired difference ($^*p<0.05$, unadjusted), and significant wins/ties/losses against the eight shared baselines (unadjusted $p<0.05$).}
\label{tab:pair-off-real}
\footnotesize
\setlength{\tabcolsep}{3pt}
\begin{tabular}{lrrrrrr}
Dataset & Budget & Acc. deployed & Acc. off & Dep.$-$Off & W/T/L deployed & W/T/L off \\
\hline
CKD & 2 & 0.9375 & 0.9875 & $-0.0500$ & 2/5/1 & 4/4/0 \\
 & 4 & 0.9875 & 0.9750 & $+0.0125$ & 2/6/0 & 1/7/0 \\
 & 7 & 0.9750 & 0.9750 & $+0.0000$ & 0/8/0 & 0/8/0 \\
ACTG175 & 3 & 0.8159 & 0.8159 & $+0.0000$ & 0/7/1 & 0/7/1 \\
 & 5 & 0.8205 & 0.8205 & $+0.0000$ & 0/6/2 & 0/6/2 \\
 & 10 & 0.8392 & 0.8392 & $+0.0000$ & 0/8/0 & 0/8/0 \\
BankMarketing & 2 & 0.7923 & 0.7923 & $+0.0000$ & 5/2/1 & 5/2/1 \\
 & 4 & 0.8034 & 0.7817 & $+0.0217^{*}$ & 7/0/1 & 4/2/2 \\
 & 7 & 0.7713 & 0.7713 & $+0.0000$ & 1/1/6 & 1/1/6 \\
PhysioNet & 5 & 0.7125 & 0.7142 & $-0.0017^{*}$ & 6/2/0 & 6/2/0 \\
 & 10 & 0.7142 & 0.7137 & $+0.0004$ & 2/5/1 & 2/5/1 \\
 & 15 & 0.7212 & 0.7212 & $+0.0000$ & 0/6/2 & 0/6/2 \\
MiniBooNE & 5 & 0.8457 & 0.8457 & $+0.0000$ & 1/1/6 & 1/1/6 \\
 & 10 & 0.8508 & 0.8636 & $-0.0128^{*}$ & 0/0/8 & 0/0/8 \\
 & 15 & 0.8699 & 0.8749 & $-0.0050^{*}$ & 0/0/8 & 0/0/8 \\
\end{tabular}
\end{table}

\textbf{Synthetic sweep (Table~\ref{tab:pair-off-alpha}).} At budget 8, removing the pair branch reverses the trend: the mean advantage falls from +0.006 at $\alpha=0$ to -0.043 at $\alpha=1$ ($r=-0.80$, negative in all five instances), against +0.007 to +0.077 ($r=+0.95$) with the branch. The branch adds 0.027, 0.051, and 0.120 accuracy at $\alpha=0.5$, 0.75 and 1 (pooled paired $p<0.0001$) and nothing detectable at $\alpha\le0.25$. At budgets 3 and 5, where the designated pair is unaffordable, removing the branch removes the losses: the advantage is positive at every $\alpha$ and within 0.005 of the feasibility-masked variant. The rise at budget 8 is therefore due to pair proposals, not to conditioning on realized values alone, and the tight-budget losses are due to the pair branch. \textbf{Fixed-$V$ control (Table~\ref{tab:pair-off-fixedv}).} Without the pair branch the advantage no longer rises with the synergy fraction ($r=-0.40$). The branch adds 0.037 and 0.025 accuracy at Syn/$V=0.90$ and 1.00, but lowers it by 0.014 and 0.012 at 0.64 and 0.78 ($p<0.001$) and is neutral at 0.51, so the benefit attributable to pairwise proposals is concentrated at high synergy fractions. \textbf{Real datasets (Table~\ref{tab:pair-off-real}).} Removing the pair branch leaves PhysioNet's win/tie/loss counts unchanged at every budget and changes accuracy by at most 0.002, so PhysioNet's low-budget advantage does not come from pair proposals. It also leaves MiniBooNE's losses in place (unchanged win/tie/loss counts at every budget), although it would raise accuracy there by 0.013 and 0.005 at budgets 10 and 15. The branch matters clearly only on BankMarketing at budget 4 (accuracy $+0.022$; wins/ties/losses 7/0/1 versus 4/2/2); the CKD differences ($n=80$) are not significant and ACTG175 is unchanged. These are test-level bootstrap comparisons on a single trained policy per dataset.

\subsection{MNIST-loop: sensitivity to the cost weight $\lambda$}
\label{app:lambda-mnist}
MNIST-loop (SynAFA v2) uses $\lambda=0.001$, whereas every tabular and synthetic experiment uses $\lambda=0.01$. The MNIST-loop value was fixed before any MNIST-loop result was inspected, because individual pixel features are sparser than tabular features and carry little information on their own: across the 784 pixels of the trained table, the median marginal information is 0.0014 bits and the 90th percentile is 0.025 bits. To measure how much this choice matters, we re-evaluated the same trained amortized table at $\lambda\in\{0.001,0.003,0.01\}$ (table fitting does not involve $\lambda$; it enters only the decision rule). This analysis was run \emph{after} the main results, on the same test set, classifier, budgets, and paired bootstrap as in Section~\ref{sec:mnist_loop} (Table~\ref{tab:lambda-mnist}).\par\smallskip\noindent

\begin{table}[htbp]
\centering
\caption{MNIST-loop (SynAFA v2) sensitivity to the cost weight $\lambda$: same trained table, only $\lambda$ changed; full 12{,}000-instance test set, shared external classifier, hard budgets 5/10/20. Accuracy of SynAFA and its difference to each baseline (accuracy(SynAFA) $-$ accuracy(baseline)); $^*$: unadjusted paired-bootstrap $p<0.05$ (10{,}000 resamples). $\lambda=0.001$ is the value used in the main text and reproduces that run exactly.}
\label{tab:lambda-mnist}
\footnotesize
\setlength{\tabcolsep}{4pt}
\begin{tabular}{llrrrr}
\multicolumn{1}{c}{\bf $\lambda$} &\multicolumn{1}{c}{\bf BUDGET} &\multicolumn{1}{c}{\bf SYNAFA ACC.} &\multicolumn{1}{c}{\bf VS. CAE} &\multicolumn{1}{c}{\bf VS. AACO} &\multicolumn{1}{c}{\bf VS. PERMUTATION} \\
\hline \\
0.001 & 5 & $0.7477$ & $-0.0235^{*}$ & $+0.1273^{*}$ & $+0.0202^{*}$ \\
      & 10 & $0.7762$ & $-0.0344^{*}$ & $+0.1491^{*}$ & $+0.0007$ \\
      & 20 & $0.8047$ & $-0.0695^{*}$ & $+0.0704^{*}$ & $-0.0182^{*}$ \\

0.003 & 5 & $0.7318$ & $-0.0394^{*}$ & $+0.1113^{*}$ & $+0.0043$ \\
      & 10 & $0.7655$ & $-0.0451^{*}$ & $+0.1384^{*}$ & $-0.0100^{*}$ \\
      & 20 & $0.8002$ & $-0.0740^{*}$ & $+0.0659^{*}$ & $-0.0227^{*}$ \\

0.01  & 5 & $0.7338$ & $-0.0374^{*}$ & $+0.1133^{*}$ & $+0.0063$ \\
      & 10 & $0.7626$ & $-0.0480^{*}$ & $+0.1355^{*}$ & $-0.0129^{*}$ \\
      & 20 & $0.7967$ & $-0.0774^{*}$ & $+0.0625^{*}$ & $-0.0261^{*}$ \\
\end{tabular}
\end{table}

\textbf{Findings.} (i) Among the three values tested, $\lambda=0.001$ gives SynAFA's highest accuracy at every budget: raising $\lambda$ lowers it by 0.45--1.59 accuracy points, each drop significant in a paired test on the same test instances. The main-text MNIST-loop numbers are therefore the most favorable to SynAFA among the tested values, and only values not smaller than 0.001 were tested. (ii) The comparisons with AACO (SynAFA better) and CAE (SynAFA worse, with a gap that widens with budget) keep their sign and significance at every $\lambda$. (iii) The comparison with permutation depends on $\lambda$: at budget 5 the advantage is $+0.0202^{*}$ at $\lambda=0.001$ but $+0.0063$ (not significant) at $\lambda=0.01$, and at budget 10 the difference moves from $+0.0007$ to $-0.0129^{*}$. (iv) Larger $\lambda$ changes which pixels are chosen: at $\lambda=0.003$ and $0.01$ the acquired set equals the $\lambda=0.001$ set in only 40--59\% of test instances. Replaying 500 test instances per budget at $\lambda=0.01$, the policy never returned a stop action, so the forced-acquisition fallback does not explain these differences. This analysis does not change the MNIST-loop conclusions (better than AACO, worse than CAE), but the size of the gap to CAE and the comparison with permutation depend on this hyperparameter.

\subsection{When does reweighting the individual PID atoms matter?}
\label{app:lambda-gating}
The main SynAFA implementation computes all four PID atoms ($R$, $U_a$, $U_b$, $\mathrm{Syn}$) but uses only their sum $V$ in the deployed pairwise score; the atoms themselves never individually influence a decision. This appendix asks the natural follow-up: if they did, would it change anything, and under what conditions?

\textbf{A controlled reweighting.} We replace $\texttt{pair\_value}[j,k]=V=R+U_a+U_b+\mathrm{Syn}$ with $Q_{\mathrm{PID}} = U_a+U_b+\gamma\,\mathrm{Syn}-\beta\,R$, recomputed from the same cached atoms, at three fixed, pre-specified settings (using $\gamma$ here to avoid collision with the synergy-fraction
sweep's interaction parameter $\alpha$ of \S\ref{sec:dose-response}): $(\gamma,\beta)\in\{(1,0),(1,1),(2,1)\}$, i.e.\ ``value minus redundancy,'' ``penalize redundancy twice,'' and ``upweight synergy specifically.'' All other scoring components of the deployed decision rule are held fixed: \texttt{single\_value} and \texttt{cond\_single\_value} (the state-dependent branch used once a pair is already started) never referenced the individual atoms and are unaffected here, so state-dependent scoring is unchanged. Only the ``propose a brand-new pair, neither member observed'' branch's ranking changes.

\textbf{The mechanism.} That branch's score is $\texttt{pair\_value}[j,k] - \lambda(\texttt{cost}_j+\texttt{cost}_k)$, compared against the best single-feature score, $\texttt{single\_value}[j]-\lambda\,\texttt{cost}_j$ (or the state-dependent conditional variant). Under the uniform per-feature costs used in these v1 datasets, the pair branch incurs twice the single-feature cost penalty. So $\lambda$ directly controls how often a decision is ever routed through the pairwise branch at all, and therefore how much room $Q_{\mathrm{PID}}$ (or any other reweighting of the same quantity) has to change anything.

\begin{figure}[h]
\centering
\includegraphics[width=0.85\linewidth]{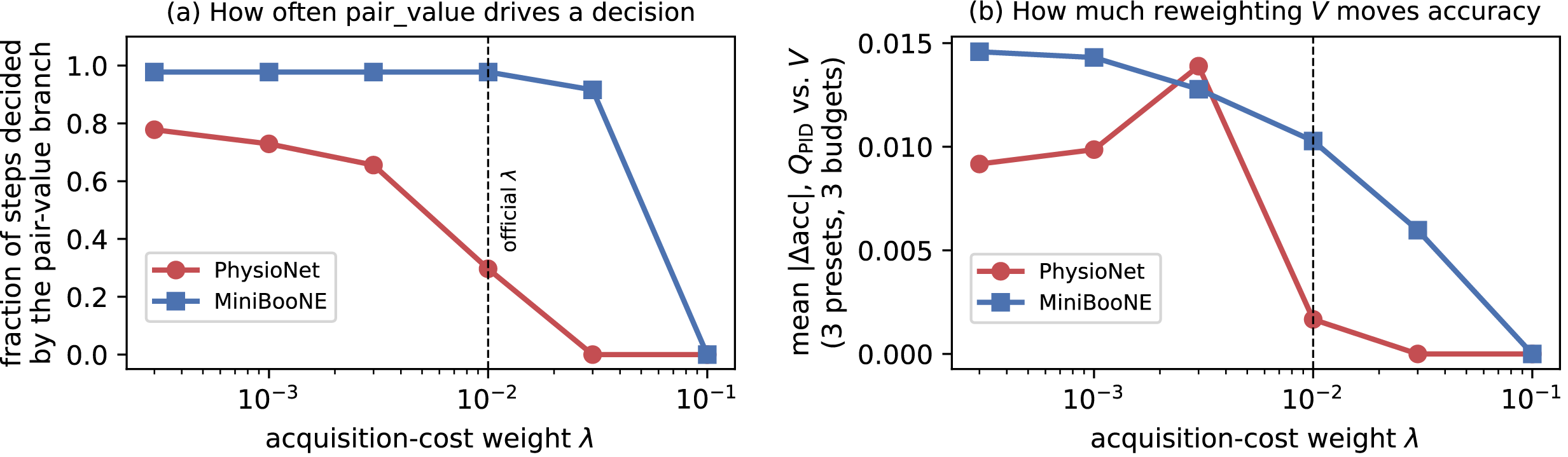}
\caption{The acquisition-cost weight $\lambda$ gates how much room PID-atom reweighting has to affect decisions (Appendix~\ref{app:lambda-gating}). (a) Fraction of acquisition steps decided by the pairwise branch of the deployed decision rule (the only branch any reweighting of pair\_value, including $Q_{\mathrm{PID}}$, can affect), as a function of $\lambda$; each point is the mean over 3 hard budgets. (b) Mean $|\Delta\mathrm{acc}|$ between $Q_{\mathrm{PID}}$ and the original joint-MI table $V$, over the same 3 presets and budgets. PhysioNet's pair-value branch, and with it $Q_{\mathrm{PID}}$'s
effect, is suppressed at a smaller $\lambda$ than MiniBooNE's (usage reaches zero at $\lambda=0.03$ versus $0.1$ on this grid); both go to exactly zero once $\lambda$ is large enough that the pair-value branch never wins.}
\label{fig:si-lambda-gating}
\end{figure}

We measured this directly: instrumenting the deployed decision rule to record which branch wins, on PhysioNet and MiniBooNE (all 3 hard budgets each), sweeping $\lambda$ from $3\times10^{-4}$ to $10^{-1}$. Figure~\ref{fig:si-lambda-gating}(a) shows the result. Averaged over the three budgets, PhysioNet's pair-value branch decides 78\% of steps at $\lambda=3\times10^{-4}$ and 30\% at the fixed $\lambda=0.01$ used in the primary experiments (45\%, 24\% and 20\% at budgets 5, 10 and 15), and exactly 0\% from $\lambda=0.03$ on: past that point, $Q_{\mathrm{PID}}$ and $V$ were reported to produce identical decisions for all three tested presets and budgets. This is an empirical observation, not a consequence of suppressing the original $V$ branch alone. MiniBooNE's pair-value branch is far more robust: 98\% of steps at every $\lambda$ up to 0.01, 92\% at $\lambda=0.03$ (100\%, 95\% and 80\% at budgets 5, 10 and 15), and 0\% at $\lambda=0.1$, one sweep point after PhysioNet's usage reaches zero. Figure~\ref{fig:si-lambda-gating}(b) shows the corresponding effect on accuracy: $Q_{\mathrm{PID}}$ vs.\ $V$'s mean $|\Delta\mathrm{acc}|$ tracks the same collapse, going to exactly zero on both datasets once their respective pair-value branch shuts off. It is not otherwise monotonic in $\lambda$: PhysioNet's effect size peaks at $\lambda=0.003$, above its value at the smallest $\lambda$ tested, before crashing.

\paragraph{Condition for invariance under reweighting.}
For $(\gamma,\beta)=(2,1)$, $Q_{\rm PID}-V=\mathrm{Syn}-2R$, which can be positive. Thus an inactive $V$ branch does not imply that every reweighted branch is inactive. Identical decisions follow if both policies remain on the unchanged single-feature branch at every visited state (with the same tie-breaking and fallback). The reported invariance must be checked for each preset, rather than inferred from original-branch frequencies alone.

At the fixed $\lambda$ used in the primary experiments (0.01 for every tabular dataset, \S\ref{sec:experiments}), the effects of PID-atom reweighting are weaker than at smaller $\lambda$ on PhysioNet (Figure~\ref{fig:si-lambda-gating}b: mean $|\Delta\mathrm{acc}|$ 0.009--0.014 for $\lambda\le0.003$ versus 0.002 at $\lambda=0.01$) and modestly weaker on MiniBooNE (0.015 versus 0.010). Repeating the full comparison across all five v1 tabular datasets,
every hard budget, and all three presets (45 dataset$\times$budget$\times$preset combinations) using the fixed $\lambda=0.01$ used in the primary evaluation: 12 of 45 show a significant difference from $V$ (8 improvements, 4 regressions), and 24 of 45 combinations show \emph{exactly zero} effect (identical decisions to $V$). Against the eight (or, for CKD and ACTG175, ten) baselines already reported in \S\ref{sec:experiments}'s tables, the aggregate win/tie/loss pattern shifts only modestly and not in a way that qualitatively rescues MiniBooNE's predominantly negative baseline comparisons: none of the 24 MiniBooNE-budget-5 comparisons (eight baselines $\times$ three presets) moves to a better outcome category than under $V$.

\textbf{Interpretation.} This is not evidence that PID-atom reweighting reliably helps or hurts; it is evidence that whether it can do either is gated by a hyperparameter this paper's primary evaluation already fixes for an unrelated reason (acquisition-cost regularization strength), and that the gate's threshold is itself dataset-dependent. A set-level scoring rule that aggregates PID information in a way calibrated against real downstream utility (Discussion, \S\ref{sec:discussion}) would need to account for this: at the budgets and cost regimes where the pairwise atoms rarely drive a decision in
the first place, no reweighting of them can be expected to matter, independent of whether the reweighting rule itself is well designed.

\end{document}